%% file: main.tex
\documentclass{article}

\usepackage[preprint]{neurips_2025}

\usepackage[utf8]{inputenc} % allow utf-8 input
\usepackage[T1]{fontenc}    % use 8-bit T1 fonts
\usepackage{url}            % simple URL typesetting
\usepackage{booktabs}       % professional-quality tables
\usepackage{amsfonts}       % blackboard math symbols
\usepackage{nicefrac}       % compact symbols for 1/2, etc.
\usepackage{microtype}      % microtypography
\usepackage{xcolor}         % colors

\usepackage{times}
\usepackage{epsfig}
\usepackage{graphicx}
\usepackage{amsmath}
\usepackage{amssymb}
\usepackage{booktabs}
\usepackage{subcaption}
\usepackage[table]{colortbl}
\definecolor{citecolor}{RGB}{30,102,235}
\usepackage[pagebackref=true,breaklinks=true,letterpaper=true,colorlinks,citecolor=citecolor,bookmarks=false]{hyperref}
\usepackage[capitalise,nameinlink]{cleveref}
\crefname{figure}{Fig.}{Figs.}
\usepackage{afterpage}

\def\method{\text{BEV-VAE}}
\title{BEV-VAE: Multi-view Image Generation with Spatial Consistency for Autonomous Driving}

\author{
Zeming Chen$\phantom{}^{1}\phantom{}$\hspace{5pt}
Hang Zhao$\phantom{}^{1,2}\phantom{}^{\text{†}}$\vspace{5pt} \\
$\phantom{}^1$Shanghai Qi Zhi Institute\hspace{5pt}
$\phantom{}^2$IIIS, Tsinghua University\hspace{5pt}
}

\begin{document}

\maketitle

{\begin{NoHyper}\let\thefootnote\relax\footnotetext{$^{\text{†}}$Corresponding to: hangzhao@mail.tsinghua.edu.cn}\end{NoHyper}}

\input{secs/0_abstract}
\input{secs/1_introduction}
\input{secs/2_related_work}
\input{secs/3_method}
\input{secs/4_experiments}
\input{secs/5_conclusion}

{\small
\bibliographystyle{unsrt}
\bibliography{egbib}
}
\clearpage
\input{supp}

\end{document}

%% file: secs/0_abstract.tex
\begin{abstract}
Multi-view image generation in autonomous driving demands consistent 3D scene understanding across camera views. 
Most existing methods treat this problem as a 2D image set generation task, lacking explicit 3D modeling. 
However, we argue that a structured representation is crucial for scene generation, especially for autonomous driving applications. 
This paper proposes BEV-VAE for consistent and controllable view synthesis. 
BEV-VAE first trains a multi-view image variational autoencoder for a compact and unified BEV latent space and then generates the scene with a latent diffusion transformer. 
BEV-VAE supports arbitrary view generation given camera configurations, and optionally 3D layouts.
Experiments on nuScenes and Argoverse 2 (AV2) show strong performance in both 3D consistent reconstruction and generation.
The code is available at~\url{https://github.com/Czm369/bev-vae}.
\end{abstract}

%% file: secs/1_introduction.tex
\section{Introduction}
Multi-view image generation is becoming increasingly important in autonomous driving, as it enables controllable synthesis of diverse scenes such as the addition or removal of vehicles based on 3D layouts. This capability facilitates the creation of rare or hard-to-collect scenarios and provides a scalable, flexible means of augmenting data for training end-to-end driving models.

Recent methods~\cite{Magicdrive, DrivingDiffusion, DriveWM, Panacea} based on fine-tuned Stable Diffusion model multi-view image generation as a set of 2D synthesis tasks with adjacent-view consistency constraints. 
While these approaches can achieve a certain degree of spatial coherence, they rely on view-dependent cross-attention in image space to implicitly model 3D structure, lacking a unified and structured scene representation. 
Consequently, they struggle to support novel view synthesis from arbitrary camera poses and cannot perform controllable generation directly conditioned on 3D layouts. 
Moreover, using 2D projections of 3D bounding boxes as conditions inevitably leads to the loss of depth information. 
Projections of different objects may overlap in image space, especially in crowded scenes, introducing occlusion ambiguity. 
As a result, the generative model must simultaneously learn to produce spatially consistent images across views and align them with these ambiguous 2D conditions, making the training process more complex and less geometrically grounded.

\begin{figure}[t]
    \centering
    \includegraphics[width=1.0\linewidth]{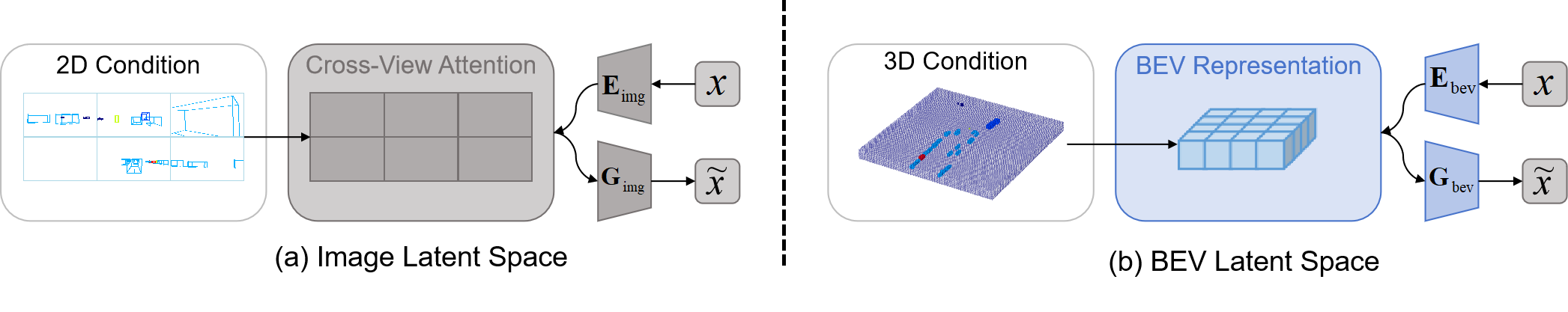}
    \vspace{-8mm}
    \caption{\textbf{Comparison of two paradigms for multi-view image generation.} (a) Image latent space generation relies on 2D projections of 3D objects to guide image synthesis and cross-view attention to enforce spatial consistency. (b) BEV latent space generation is conditioned on 3D occupancy to produce a unified representation, from which all views are decoded, naturally preserving spatial consistency and enabling novel view synthesis by adjusting camera poses.}
    \label{fig:paradigm}
    \vspace{-6mm}
\end{figure}

In contrast, our approach adopts a fundamentally different paradigm by performing generation in a Bird’s-Eye-View (BEV) latent space, as shown in Fig.~\ref{fig:paradigm}. 
Instead of modeling each view separately, BEV-VAE encodes a unified latent representation that captures both semantic content and structured 3D spatial geometry. 
This shared BEV representation ensures spatial consistency across all views, as the same spatial location corresponds to consistent content regardless of camera perspective. Novel views can be synthesized simply by modifying camera poses at decoding time, without the need for retraining.
Furthermore, object layouts can be explicitly edited using 3D binary occupancy maps, which are spatially aligned with the BEV latent space. 
This alignment enables precise and interpretable control over object quantity, position, and category, and avoids the ambiguity and lack of depth information introduced by 2D projections of 3D bounding boxes.

In this paper, we propose BEV-VAE, a multi-view image generation method with a unified representation of the 3D scene. BEV-VAE explicitly constructs a spatially aligned latent space in bird's-eye view (BEV) during the encoding stage,  This structured BEV space enables high-fidelity reconstruction with strong cross-view alignment, supports novel view synthesis by manipulating camera poses without retraining, and allows controllable generation conditioned on 3D object layouts, such as varying object quantity, position, or category—offering a more scalable and interpretable solution for autonomous driving applications.
Experiments on nuScenes and Argoverse 2 (AV2) show strong reconstruction and generation performance. BEV-VAE is the first to generate all 7 surround-view images on AV2, demonstrating its robustness and practicality.

Our contributions are as follows.

$\bullet$ We propose a framework that constructs spatially aligned BEV representations from multi-view images, enabling high-fidelity reconstruction with strong cross-view consistency.

$\bullet$ We demonstrate that the learned BEV latent space supports novel view synthesis by manipulating camera poses, validating its structured 3D nature and spatial coherence.

$\bullet$ We instantiate diffusion-based generation directly in the BEV space, allowing controllable synthesis conditioned on 3D object layouts, such as quantity, location and category.

%% file: secs/2_related_work.tex
\section{ Related Work}
\subsection{Bird's-Eye-View Perception}
Autonomous driving relies on Bird's Eye View (BEV) to unify multi-view image information. 
The construction of the BEV feature follows two approaches: bottom-up and top-down. 
Bottom-up methods~\cite{LSS, BEVDet, BEVFusion} estimate the depth required to lift 2D features into 3D space before fusing them into BEV. 
In contrast, top-down methods~\cite{BEVFormer, UniAD} use deformable attention and query mechanisms to efficiently aggregate features by dynamically sampling key regions.

In top-down methods, deformable attention (DA) plays a pivotal role in enhancing computational efficiency and focusing on relevant areas. Let $q, p$, and $v$ represent the query, reference points, and value features, respectively. $M$ denotes the number of attention heads and $K$ is the total number of sampled keys. The mechanism is calculated by:
$\operatorname{DA}(q, p, v)=\sum_{m=1}^{M} \mathcal{W}_m \sum_{k=1}^{K} \mathcal{A}_{m k} \cdot \mathcal{V}_{m k},$
where $m$ indexes the attention head, and $k$ indexes the sampled keys. 
The $\mathcal{W}_m \in \mathbb{R}^{C \times C / M}$ are learnable weights with dimension $C$, and $\mathcal{V}_{m k}$ are the features at location $p + \Delta p_{m k}$, which are extracted by bilinear interpolation.
$\Delta p_{m k}$ and $\mathcal{A}_{m k}$ denote the sampling offset and attention weight of the $k^\text{th}$ sampling point in the $m^\text{th}$ attention head, respectively.
Both $\Delta p_{m k}$ and $\mathcal{A}_{m k}$ are obtained via linear projection over the query $q$, and $A_{m k}$ is normalized by softmax to ensure $\sum_{k=1}^{K} \mathcal{A}_{m k}\!= \!1$.

\subsection{Variational Autoencoder}
Variational AutoEncoder (VAE) formulates image generation as probabilistic inference by introducing a latent variable $z$ and optimizing the Evidence Lower Bound (ELBO) to jointly learn a Gaussian posterior and a reconstruction decoder.
However, modeling $z$ as a Gaussian limits the sharpness and fidelity of generated images.
To address this, VQVAE~\cite{VQVAE} introduces a discrete codebook to enhance diversity, while VQGAN~\cite{VQGAN} integrates GAN~\cite{GAN} training with VQVAE, leveraging adversarial, perceptual~\cite{PerceptualLoss}, and reconstruction losses for realistic image generation. 
ViT-VQGAN~\cite{ViT-VQGAN} further improves global context modeling and codebook efficiency by employing ViT~\cite{ViT} as both encoder and decoder, enhancing generative performance.
In addition to modeling the latent variable $z$ as a discrete distribution, diffusion models transform data into a standard Gaussian distribution by progressively adding noise and then reverse this process via denoising. 
DDPM~\cite{DDPM} employs a Markov chain for iterative noise addition and recovery, while DDIM~\cite{DDIM} accelerates sampling with deterministic inference. 
LDM~\cite{LDM} performs diffusion in a compressed latent space to enhance efficiency and quality. 
DiT~\cite{DiT} further integrates diffusion with Transformer architectures, improving high-resolution generation and expanding applicability.

\subsection{Autonomous Driving Multi-view Generation}
Multi-view generation inherently represents a 3D scene through 2D images.
BEVGen~\cite{BEVGen} uses autoregressive generation to produce multi-view images based on BEV layouts. It constructs direction vectors for cameras and BEV layouts, maps them to the BEV ego-vehicle coordinate system via camera parameters, and integrates their inner product as an attention bias to enhance spatial consistency.
However, subsequent works have increasingly adopted diffusion-based generation methods, fine-tuning Stable Diffusion to transfer its conditional generation capabilities to the autonomous driving domain.
DrivingDiffusion~\cite{DrivingDiffusion}, Magicdrive~\cite{Magicdrive}, and Panacea~\cite{Panacea} utilize cross-attention on adjacent view images to ensure consistency between perspectives.
MagicDrive integrates camera pose information by encoding camera parameters similar to NeRF~\cite{NeRF}, while Panacea extends this approach by generating pseudo-RGB images of camera frustum directions and embedding pose information through ControlNet~\cite{ControlNet}.
Additionally, DriveWM~\cite{DriveWM} uses self-attention to fuse spatially aligned features across views and predicts stitched views between nonadjacent references to maintain multi-view spatial consistency.

%% file: secs/3_method.tex
\section{Method}
\begin{figure}[t]
    \centering
    \includegraphics[width=1.0\linewidth]{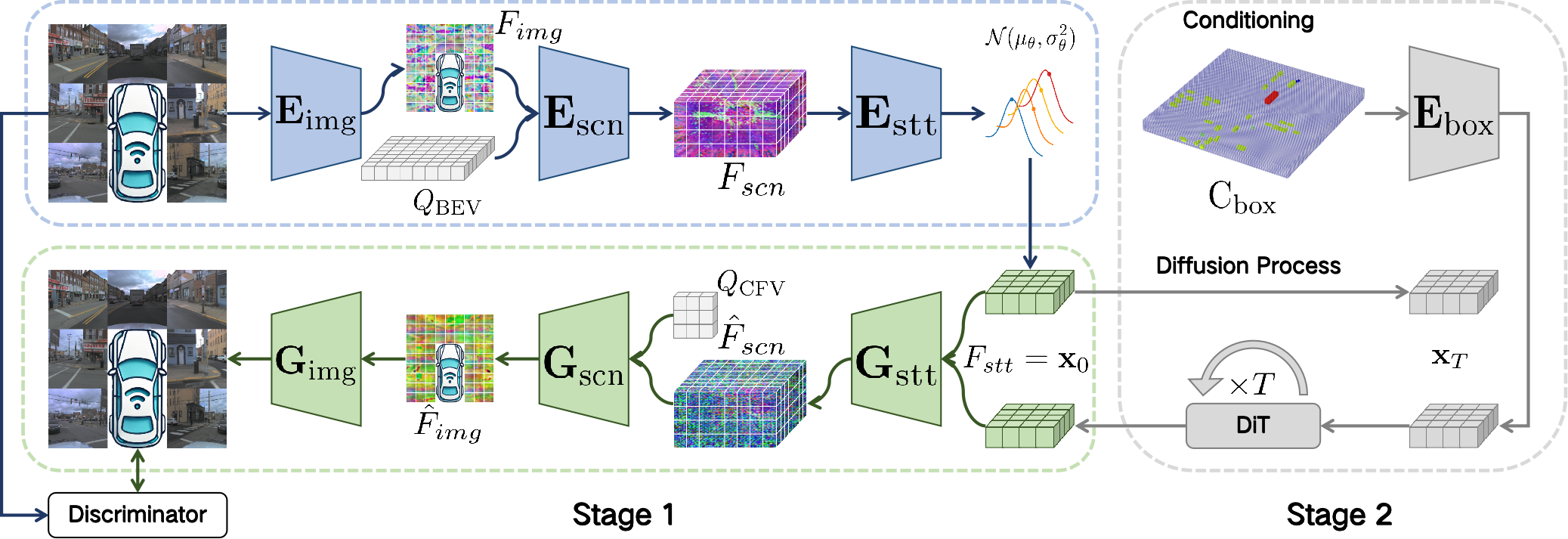}
    \vspace{-4mm}
    \caption{\textbf{Overall architecture of \method~with DiT for multi-view image generation.} In Stage 1, BEV-VAE learns to encode multi-view images into a spatially compact latent space in BEV and reconstruct them, ensuring spatial consistency. In Stage 2, DiT is trained with Classifier-Free Guidance (CFG) in this latent space to generate BEV representations from random noise, which are then decoded into multi-view images.}
    \label{fig:framework}
    \vspace{-6mm}
\end{figure}
\subsection{Overall Architecture}
\method~consists of a Transformer-based encoder $E$, decoder $G$, and a StyleGAN discriminator $D$.
The encoder $E$ maps multi-view images into a latent Gaussian distribution via its image, scene, and state encoders, from which state features are sampled via reparameterization. 
The decoder $G$, comprising state, scene, and image decoders, reconstructs spatially consistent multi-view images from the state features, ensuring spatial consistency across views. 
The discriminator $D$ distinguishes real from reconstructed images, guiding $G$ with adversarial loss.
Both encoder $E$ and decoder $G$ are trained with KL divergence, reconstruction, and adversarial losses.
Additionally, a DiT performs denoising in the BEV latent space, enabling multi-view image generation.
\subsection{Encoder}
\noindent\textbf{Image Encoder} employs ViT with a patch size of $8$ to encode a $256 \times 256$ image into a $32 \times 32$ token sequence. To capture semantic information and local details for 3D scene encoding, an upsampling-only FPN~\cite{FPN} constructs a three-level feature pyramid to enhance multi-scale representation. The process can be formulated as:
$
F_{img} = \operatorname{FPN}(\mathbf{E}_{\mathrm{img}}(x)) = \operatorname{Concat}(F_{img}^0, F_{img}^1, F_{img}^2),
$
where $F_{img}^i \in \mathbb{R}^{V \times L_i \times C}(i \in[0,2])$ are the multi-scale flattened image features with $C=96$ and sequence length $L_i=32 \times 32 \times 2^{2i}$. 
Here, $V$ is the number of views.

\noindent\textbf{Scene Encoder} utilizes a deformable attention mechanism to construct 3D scene features by extracting multiview image features. 
A $128 \times 128$ grid of pillars is pre-defined around the ego vehicle in BEV, each with a height of 8. 
All reference points in the same pillar share a learnable query, while different height positions are distinguished through positional encoding.
The reference points of scene features are projected onto image features by camera parameters, enabling BEV queries to aggregate spatially aligned features from multiview image features via deformable attention.
The process can be formulated as:
$
F_{scn}=\frac{1}{|\mathcal{V}_{\text {hit}}|} \sum_{v \in \mathcal{V}_{\text {hit }}} \operatorname{DA}(Q_{\text{BEV}}, P_{\text{BEV}}, F_{img}^{(v)}),
$
where $Q_{\text{BEV}} \in \mathbb{R}^{L_{Q} \times C}$ are the flattened 3D BEV queries with $C=96$, $P_{\text{BEV}} \in \mathbb{R}^{L_{Q} \times 3}$ denote the corresponding reference points, $F_{img}^{(v)} \in \mathbb{R}^{L_{V} \times C}$ is the image feature sequence of the view $v$, and the set $\mathcal{V}_{\text {hit}}$ refers to the views containing projected reference points, ensuring that only relevant views contribute to the aggregated scene feature.
Here, $L_Q=8 \times 128 \times 128$ is the BEV query sequence length, and $L_V=\sum_{i=0}^2(32 \times 32 \times 2^{2i})$ is the total image feature sequence length across resolutions.

\noindent\textbf{State Encoder} integrates multi-height scene features in BEV by concatenating them along the height dimension, reshaping the input from $96 \times 8 \times 128 \times 128$ to $768 \times 128 \times 128$.
It then partitions the features into $32 \times 32$ patches along the horizontal plane, reducing the computational cost while introducing local receptive fields.
Finally, it applies self-attention to model global spatial relationships and encode highly compressed spatial state features.

\subsection{Decoder}
\noindent\textbf{State Decoder} is responsible for reconstructing structurally detailed 3D scene features from the compressed 2D state representation. 
It first applies self-attention to capture global spatial relationships, and then regroups the features to restore horizontal and height structures. 
The state features are first expanded from $32 \times 32$ to $128 \times 128 $ along the horizontal plane through deconvolution, then further transformed from $768 \times 128 \times 128$ to the original multi-height format $96 \times$ $8 \times 128 \times 128$ through dimension partitioning.
To refine 3D scene feature decoding, a downsampling-only FPN is employed, effectively reconstructing detailed structures across scales. The process can be formulated as:
$
\hat{F}_{scn} = \operatorname{FPN}(\mathbf{G}_{\mathrm{stt}}(\hat{x})) = \operatorname{Concat}(\hat{F}_{scn}^0, \hat{F}_{scn}^1, \hat{F}_{scn}^2),
$
where $\hat{F}_{s c n}^i \in \mathbb{R}^{L_i \times C}(i \in[0,2])$ are the reconstructed multi-scale flattened scene features with $C=96$ and sequence length $L_i=8 \times 128 \times 128 \times 2^{-3i}$.

\noindent\textbf{Scene Decoder} transforms scene features from the Bird's Eye View (BEV) to the Camera's Frustum View (CFV) and aggregates multi-depth information to reconstruct image features.
A $32 \times 32$ frustum of rays is predefined per camera, each spanning 60 depth levels.
All reference points along the same ray share a learnable query, while different depth positions are distinguished through positional encoding.
Similar to the projection of reference points of scene features from BEV onto image features via camera parameters, reference points of scene features in CFV can also be projected to BEV, enabling CFV queries to construct features along depth dimensions for different views via deformable attention.
Furthermore, CFV queries estimate depth weights to perform a weighted summation of the features at all reference points along the ray, thereby generating the projected image features.
Considering that some reference points may exceed the range of scene features, their corresponding weights are set to 0. 
The process can be formulated as:
$
\hat{F}^{(v)}_{img} = \sum_{d \in \mathcal{D}_{\text{hit}}}W_{d} \odot \operatorname{DA}(Q_{\text{CFV}}, P_{\text{CFV}}, \hat{F}_{scn}),
$
where $Q_{\text{CFV}} \in \mathbb{R}^{L_{Q} \times C}$ are the flattened 3D CFV queries with $C=96$, $P_{\text{CFV}} \in \mathbb{R}^{L_{Q} \times 3}$ denote the corresponding reference points, $\hat{F}_{scn} \in \mathbb{R}^{L_{V} \times C}$ is the reconstructed scene feature sequence, and the set $\mathcal{D}_{\text {hit }}$ refers to the depth positions along the ray where reference points fall within the valid scene feature range, ensuring that only effective depth positions contribute to the aggregated image feature.
Here, $L_Q=60 \times 32 \times 32$ is the CFV query sequence length, and $L_V=\sum_{i=0}^2(8 \times 128 \times 128 \times 2^{-3i})$ is the total reconstructed scene feature sequence length across resolutions.

\noindent\textbf{Image Decoder} progressively restores pixel-level details by processing scene features projected onto the image plane.
As its preceding stage, the scene decoder aggregates scene features along the ray depth dimension but lacks interactions between rays. 
To complement this, it maps the projected scene features $(C=96)$ to 768 dimensions via a linear layer, models global spatial and semantic relationships on the image plane by self-attention, and upscales the resolution from $32 \times 32$ to $256 \times 256$ with deconvolution, reconstructing fine-grained image details.

\subsection{Loss}
\noindent\textbf{KL Divergence Loss} regularizes the latent distribution of the state features, enforcing closeness to a standard normal distribution and ensuring continuity in the latent space:
$
\mathcal{L}_{\mathrm{KL}} = D_{\mathrm{KL}}(q_\phi(z \mid x) \| p(z)) = \frac{1}{2} \sum_{i=1}^d(\sigma_i^2+\mu_i^2-1-\log \sigma_i^2),
$
where $p(z)$ is defined as $\mathcal{N}(0, I)$, $d$ is the dimension of state features, and $\mu_i, \sigma_i^2$ are the mean and variance of the $i$-th latent dimension predicted by the encoder $E$.
To allow gradient-based optimization of the stochastic sampling process, the reparameterization trick is used. Instead of directly sampling $z$ from $q_\phi(z \mid x)$, it is reparameterized as:
$
z=\mu+\sigma \odot \epsilon,\quad(\mu,\sigma)=E(x),\quad \epsilon \sim \mathcal{N}(0, I).
$

\noindent\textbf{Reconstruction Loss} ensures that the reconstructed image $\hat{x}=G(z)$ retains both pixel-level details and high-level semantic structure of the target image $x$. This is achieved by combining pixel-wise loss with perceptual loss:
$
\mathcal{L}_{\mathrm{R}} = \mathcal{L}_{2} + \mathcal{L}_{\mathrm{perceptual}} = \|x-\hat{x}\|^2 + \sum_{l} \|\psi_{l}(x)-\psi_{l}(\hat{x})\|^2.
$
Here, $\mathcal{L}_2$ enforces pixel-wise similarity between the image $x$ and its reconstruction $\hat{x}$, while $\mathcal{L}_{\text {perceptual }}$ captures structural and semantic consistency by comparing feature maps $\psi_l(x)$ and $\psi_l(\hat{x})$ extracted from the $l$-th layer of a pre-trained VGG-16.
This balance preserves fine details and perceptual coherence, yielding realistic reconstructions.

\noindent\textbf{Discriminator Loss} enables the discriminator $D$ to distinguish real images from reconstructed ones, improving its ability to provide meaningful adversarial feedback. With the hinge loss formulation, it is expressed as:
$
\mathcal{L}_{\mathrm{D}}=\max (0,1-D(x))+\max (0,1+D(\hat{x})),
$
which encourages the discriminator to assign higher scores to real images and lower scores to reconstructed ones. Hinge loss stabilizes adversarial training by preventing excessively large gradients for confident predictions while ensuring effective feedback for refining reconstruction quality, leading to more stable and efficient optimization.

\noindent\textbf{Adversarial Loss} leverages the discriminator’s feedback to enhance the perceptual realism of reconstructed images and is defined as:
$
\mathcal{L}_{\mathrm{A}}=-D(\hat{x})
$

\noindent\textbf{Total Loss for Encoder and Decoder} combines the KL divergence loss, reconstruction loss, and adversarial loss, ensuring effective latent space regularization and perceptual realism. It is formulated as:
$
\mathcal{L}_{\mathrm{G}}=\beta \cdot \mathcal{L}_{\mathrm{KL}}+ \mathcal{L}_{\mathrm{R}} +0.1 \cdot \lambda \cdot\mathcal{L}_{\mathrm{A}}
$
where $\beta=10^{-6}$ controls the strength of the KL divergence regularization. 
The adaptive weight $\lambda$ balances the adversarial loss relative to the reconstruction loss, ensuring that the adversarial term contributes meaningfully without overpowering reconstruction. It is computed as
$
\lambda=\frac{\nabla_{G_L}[\mathcal{L}_{\mathrm{R}}]}{\nabla{G_L}[\mathcal{L}_{\mathrm{A}}]+\delta} 
$
with $\nabla{G_L}[\cdot]$ denoting the gradient of the corresponding term with respect to the last layer $L$ of the decoder, and $\delta=10^{-6}$ ensuring numerical stability.

\subsection{Generation}
\noindent\textbf{BEV-VAE w/ DiT} extends BEV-VAE by integrating DiT in its latent space, leveraging CFG to enhance conditional generation. 
By explicitly incorporating structured occupancy constraints from 3D object bounding boxes, it ensures spatial consistency and controllability in generation.
Given a set of 3D bounding boxes $\{\mathbf{b}_i\}_{i=1}^N$, each parameterized as:
$
\mathbf{b}=(q_w, q_x, q_y, q_z, x_c, y_c, z_c, l, w, h, c),
$
where the quaternion $q=(q_w, q_x, q_y, q_z)$ encodes the 3D orientation, $(x_c, y_c, z_c)$ specifies the box center in the ego coordinate system, $(l, w, h)$ represents the size of the box, and $c \in {1, \ldots, C}$ is the semantic class index.
These boxes are voxelized into a binary occupancy tensor $\mathbf{C}_{\text{box}} \in \{0,1\}^{C \times 8 \times 128 \times 128}$, where each voxel represents whether a given spatial location is occupied by a bounding box of a particular class. Formally, it is defined as: 
$
\mathbf{C}_{\mathrm{box}}(c, z, y, x)=\max _{i: c_i=c} \mathbf{1}[(z, y, x) \in \Omega(\mathbf{b}_i)]
$
where $\mathbf{1}[\cdot]$ is an indicator function, and $\Omega(\mathbf{b}_i)$ denotes the discretized voxelized representation of bounding box $\mathbf{b}_i$. The max operation aggregates occupancy information from overlapping bounding boxes within the same class.
The occupancy tensor $\mathbf{C}_{\mathrm{box}}$ is downsampled via non-overlapping patch partitioning in the BEV plane, yielding a feature of shape $96 \times 8 \times 32 \times 32$, followed by channel-wise concatenation of the height dimension to form the conditional occupancy feature $F_{{box}} \in \mathbb{R}^{768 \times 32 \times 32}$. Aligned with the state feature $F_{\text {stt, }}$, it is injected via element-wise addition:
$
F_{stt}^{\prime} = F_{stt} + s \cdot F_{box}, 
$ 
where $s$ is the guidance scale in CFG.
This ensures spatial consistency by aligning the conditional occupancy features and state features within the shared BEV coordinate system, allowing DiT to focus on relevant regions by explicitly incorporating object category and location information.

%% file: secs/4_experiments.tex
\section{Experiments}
\begin{figure}[t]
    \centering
    \includegraphics[width=1.0\linewidth]{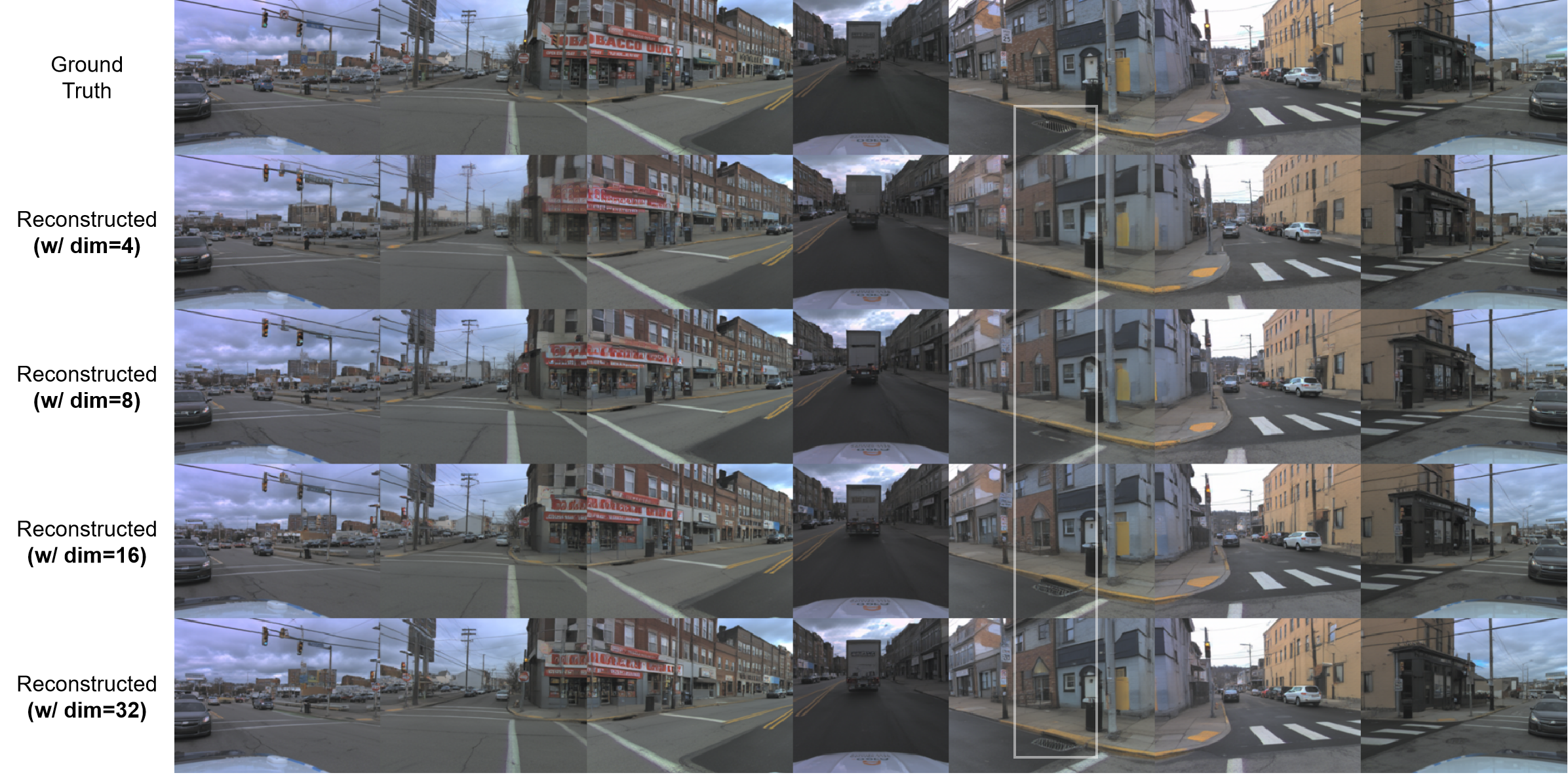}
    \vspace{-4mm}
     \caption{\textbf{Multi-view image reconstruction on AV2.} Row 1 shows validation images, and Rows 2-5 display reconstructed images with latent dimensions of 4, 8, 16, and 32. With higher dimensions, the reconstruction more accurately preserves fine details, such as the manhole covers in the white box.}
    \label{fig:rec_dim}
    \vspace{-4mm}
\end{figure}
\begin{figure}[t]
    \centering
    \includegraphics[width=1.0\linewidth]{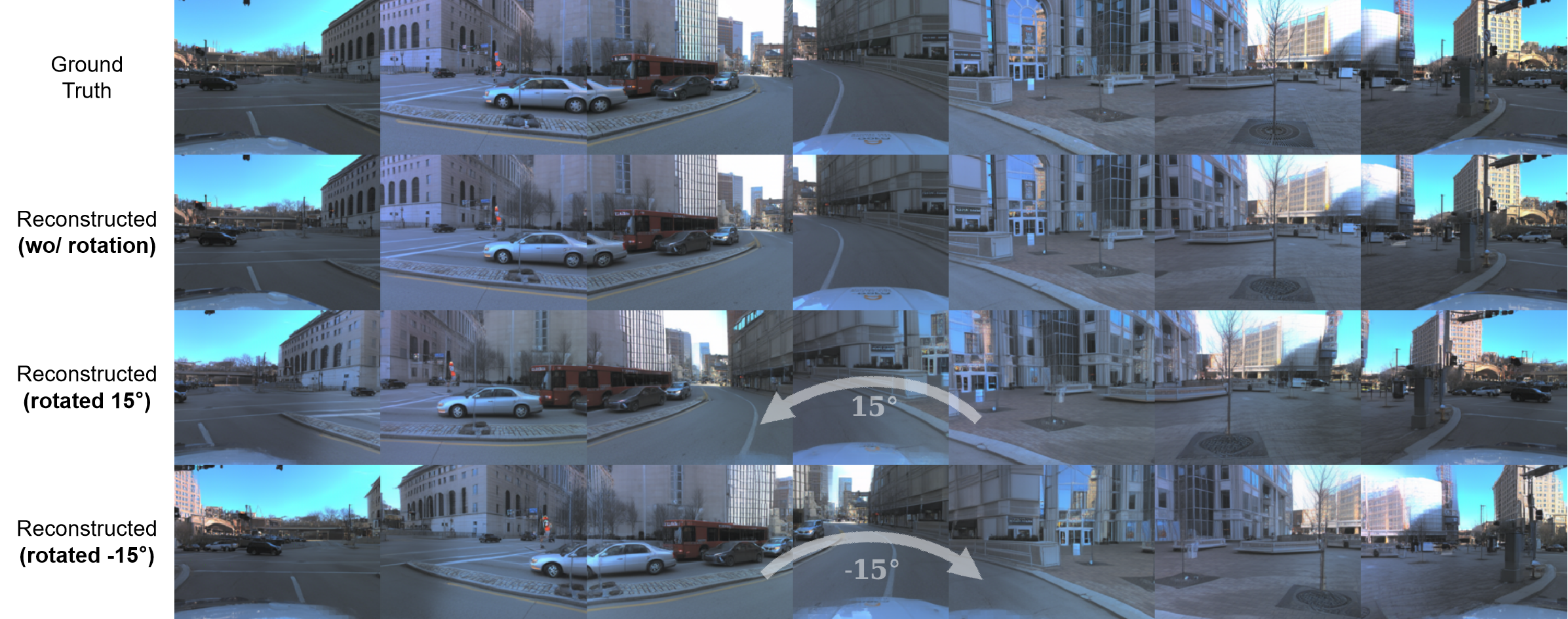}
    \vspace{-4mm}
    \caption{\textbf{Novel view synthesis via camera pose modifications.} Row 1 presents validation images, and Row 2 shows reconstructions. Rows 3 and 4 display reconstructed images with all cameras rotated 15° left and 15° right, respectively. Note: Latent dimension is set to 32.}
    \label{fig:rec_rot_ego}
    \vspace{-6mm}
\end{figure}
\begin{figure}[t]
    \centering
    \includegraphics[width=1.0\linewidth]{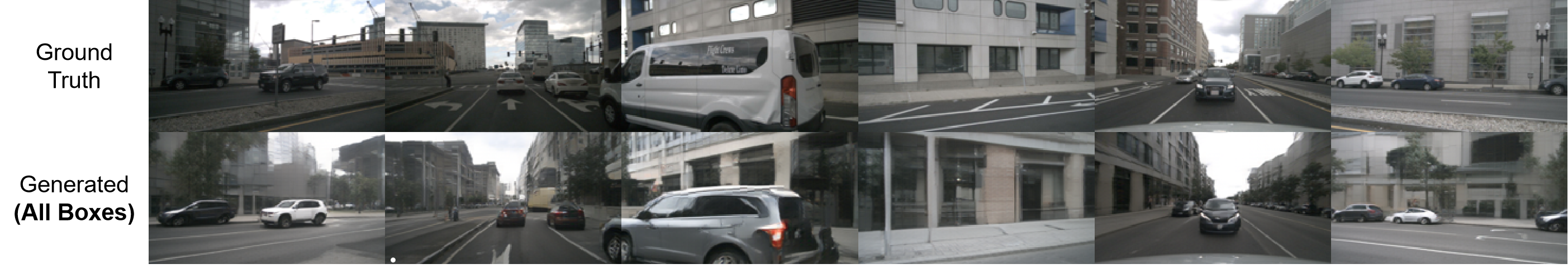}
    \vspace{-4mm}
    \caption{\textbf{Multi-view image generation on nuScenes.} Row 1 shows validation images, and Row 2 presents images generated from the corresponding 3D bounding boxes. }
    \label{fig:gen_nusc}
    \vspace{-4mm}
\end{figure}
\begin{figure}[t]
    \centering
    \includegraphics[width=1.0\linewidth]{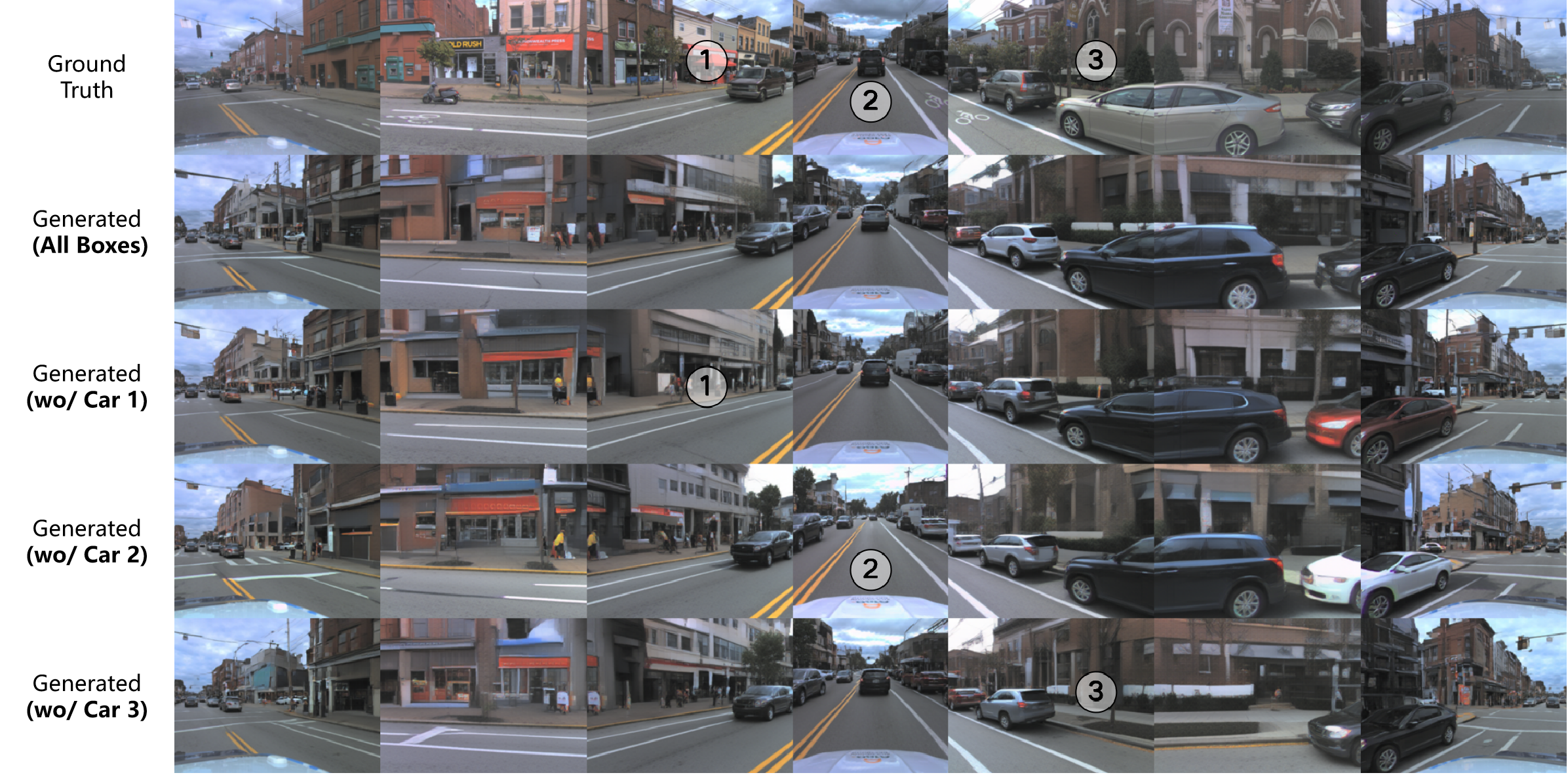}
    \vspace{-4mm}
    \caption{\textbf{Multi-view image generation on AV2.} Row 1 presents real images from the validation set. Row 2 shows images generated from the corresponding 3D bounding boxes. Rows 3-5 present generated images after removing a specific vehicle, with the removed vehicles indicated by numerical labels. Note: The same 3D bounding box may produce different objects across generated images.}
    \label{fig:gen_drop}
    \vspace{-6mm}
\end{figure}
\subsection{Datasets}
This study uses two multi-camera datasets, nuScenes and Argoverse 2 (AV2), which provide synchronized multi-camera images and 3D object bounding boxes.

\noindent\textbf{The nuScenes dataset} consists of 6 cameras with 700 training scenes and 150 validation scenes. Each scene contains approximately 220 samples, of which 40 are annotated across 10 object categories. In total, it includes 155,245 training samples, of which 28,126 are annotated, and 33,142 validation samples, of which 6,019 are annotated.

\noindent\textbf{The AV2 dataset} consists of 7 cameras, with the front camera rotated by 90°. It includes 700 training scenes and 150 validation scenes. Each scene contains approximately 300 samples, of which 150 are annotated across 30 object categories. In total, it includes 224,175 training samples, of which 109,907 are annotated, and 47,946 validation samples, of which 23,521 are annotated.

\subsection{Metrics}
The performance of BEV-VAE is evaluated using multiple metrics covering reconstruction quality, multi-view spatial consistency, and generation quality.

\noindent\textbf{PSNR and SSIM} measure the similarity between reconstructed and original images, with PSNR assessing signal fidelity and SSIM focusing on structural consistency.

\noindent\textbf{Multi-View Spatial Consistency (MVSC)} evaluates spatial consistency in multi-view reconstruction. 
Following BEVGen~\cite{BEVGen} and DriveWM~\cite{DriveWM}, a pre-trained LoFTR~\cite{LoFTR} is used to compute keypoint matching confidence between adjacent views.
MVSC is the ratio of average adjacent-view matching confidence in reconstructed images to that in real images, where higher values imply better alignment.

\noindent\textbf{FID} quantifies the distributional difference between original and target images in a deep feature space. It is used to evaluate both reconstruction quality and the quality of generated multi-view images.

\subsection{Settings}
All experiments are conducted on a single machine with 8 NVIDIA A800 GPUs.
The training process consists of two stages, both utilizing the AdamW optimizer.

\noindent\textbf{Stage 1:} The batch size is set to 1 per GPU, with a learning rate of 4.0e-5 for nuScenes and 8.0e-5 for AV2. Training lasts for 100k iterations with a 5k warm-up, betas (0.9, 0.99), weight decay 1e-4, and EMA decay 0.9999.

\noindent\textbf{Stage 2:} The batch size is set to 8 per GPU, with a learning rate of 1.0e-4. Training spans 400k iterations for nuScenes and 200k for AV2, with a 5k warm-up, betas (0.9, 0.95), weight decay 0.1, bias decay 0.0, and EMA decay 0.999.

\subsection{Reconstruction}
\begin{table}[t]
\centering
\vspace{-4mm}
\input{tabs/rec_all}
% \vspace{-6pt}
\caption{\textbf{Comparison of BEV-VAE with varying latent dimensions and SD-VAE for multi-view reconstruction.}
BEV-VAE performs spatial modeling by encoding multi-view images into a unified BEV representation and decoding them back into images, while SD-VAE, trained on 5.85 billion images, serves only as a reference for image reconstruction rather than a direct baseline.}
\label{tab:rec_all}
\vspace{-8mm}
\end{table}
This section investigates the latent dimensionality $D$ required for BEV-VAE to encode multi-view images into a unified BEV representation that preserves 3D structure and semantics. Unlike SD VAE, which compresses a single $256\times256$ image into a $32\times32\times4$ representation, BEV-VAE encodes multiple camera views into a shared $32\times32\times D$ representation, capturing richer spatial and semantic information. As this representation integrates multiple views and encodes spatial structure, it is evident that $D>4$ is necessary.
We analyze how varying $D$ affects reconstruction quality on nuScenes and AV2, as summarized in Tab.~\ref{tab:rec_all}. SD-VAE (i.e., the AutoencoderKL of Stable Diffusion) is included as a reference standard rather than a baseline. Trained on 5.85 billion images from LAION-5B~\cite{Laion-5b}, SD-VAE achieves exceptional image fidelity and serves as a strong latent backbone for generative modeling. In contrast, BEV-VAE is trained on much smaller-scale multi-view datasets (155K samples for nuScenes and 224K for AV2), despite its strong spatial modeling capabilities.
Despite the data scale gap, BEV-VAE demonstrates strong performance. As shown in Tab.~\ref{tab:rec_all}, increasing the latent dimension $D$ consistently improves reconstruction quality. However, in nuScenes, FID slightly worsens at $D=32$, likely due to overfitting. In contrast, AV2, which contains 1.5× more samples, continues to benefit from increased capacity, suggesting improved generalization.
Notably, while BEV-VAE still lags behind SD VAE in PSNR, SSIM, and FID, it outperforms SD-VAE in MVSC (Multi-View Spatial Consistency) on AV2. This indicates that BEV-VAE decouples spatial consistency from single-view image quality: since all views are decoded from the same BEV representation, overlapping regions across views inherently share features from identical spatial locations in the latent space.
Fig.\ref{fig:rec_dim} provides a qualitative comparison: smaller $D$ values lead to blurry reconstructions and misaligned views, while larger values yield better fidelity and alignment. Furthermore, BEV-VAE enables novel view synthesis by modifying camera poses, as shown in Fig.\ref{fig:rec_rot_ego}, demonstrating spatially consistent generations from unseen viewpoints.

\subsection{Generation}
\begin{table}[t]
\vspace{-4mm}
\begin{minipage}[t]{0.46\textwidth}
\setlength{\tabcolsep}{9.0pt}
\scalebox{0.8}{\input{tabs/gen_nusc}}
\centering
\vspace{-6pt}
\subcaption{FID on nuScenes across scales and dimensions.}
\label{tab:gen_nusc}
\end{minipage}
\hspace{0.06\textwidth}
%\hfill
\begin{minipage}[t]{0.46\textwidth}
\setlength{\tabcolsep}{9.0pt}
\scalebox{0.8}{\input{tabs/gen_av2}}
\centering
\vspace{-6pt}
\subcaption{FID on AV2 across scales and dimensions.}
\label{tab:gen_av2}
\end{minipage}
\vspace{-6pt}
\centering
\caption{\textbf{Impact of guidance scale across latent dimensions for multi-view image generation}.}
\vspace{-8mm}
\end{table}

The impact of different guidance scale ($s$) on the generation quality of DiT trained with CFG in various latent space dimensions is analyzed. 
Increasing the latent space dimension improves reconstruction quality but also makes the generation task more challenging to learn.
Table \ref{tab:gen_nusc} and Table \ref{tab:gen_av2} present the experimental results on nuScenes and AV2, respectively. 
In both cases, the optimal latent space dimension is $dim=8$. 
The highest generation fidelity is achieved at $s=5$ on nuScenes (FID = 21.14) and at $s=3$ on AV2 (FID = 10.68). 
The lower FID on AV2 suggests that its larger scale and greater diversity provide richer training signals, leading to improved generation quality.
Since the conditioning matrix in CFG is spatially aligned with the latent variables of BEV-VAE, it enables explicit control over both the quantity and position of objects.
As shown in Fig.~\ref{fig:gen_drop}, the generation results on AV2 demonstrate that specific vehicles can be selectively removed, allowing direct comparison with real images.
Table~\ref{tab:benchmark} compares BEV-VAE with previous multi-view image generation methods on nuScenes. 
Both BEV-VAE and BEVGen are trained from scratch without pre-trained priors, making their comparison fair. 
BEV-VAE achieves significantly better generation quality than BEVGen, demonstrating the effectiveness of modeling multi-view generation as a 3D scene generation task to enhance spatial consistency. 
Additionally, BEV-VAE is applicable to autonomous driving datasets with varying numbers of cameras, highlighting its broad adaptability. 
While methods fine-tuned from Stable Diffusion still perform better, BEV-VAE exhibits strong scalability, as its performance continues to improve with increasing training data, making it a promising approach for large-scale multi-view generation.
\begin{table}[h]
\vspace{-4mm}
\centering
\begin{minipage}{0.46\linewidth}
\setlength{\tabcolsep}{6.0pt}
\scalebox{0.7}{\input{tabs/benchmark}}
% \vspace{-6pt}
\vspace{1pt}
\caption{\textbf{Benchmark results on nuScenes.} BEV‑VAE w/ DiT significantly bridges the gap from-scratch and SD-finetuned methods.}
\label{tab:benchmark}
\end{minipage}
\hspace{0.06\textwidth}
\begin{minipage}{0.46\linewidth}
\setlength{\tabcolsep}{6.0pt}
\scalebox{0.7}{\input{tabs/ablation}}
\centering
% \vspace{-6pt}
\caption{\textbf{Ablation on loss function for reconstruction performance on AV2.} The latent dimension is fixed at 8.}
\label{tab:ablation}
\end{minipage}
\vspace{-8mm}
\end{table}
\subsection{Analysis of Loss Function}
Table~\ref{tab:ablation} presents the ablation study results for different loss configurations in BEV-VAE. 
Removing the perceptual loss $\mathcal{L}_{\mathrm{perceptual}}$ causes a sharp FID increase from 5.10 to 68.99, highlighting its critical role in enhancing perceptual quality. 
Although PSNR and SSIM remain relatively stable, the degradation in FID suggests a loss of fine details and realism.
The adversarial loss $\mathcal{L}_{\mathrm{A}}$ significantly impacts realism, as its removal increases FID to 13.62. 
Interestingly, MVSC slightly improves, indicating that adversarial training refines high-frequency details but may introduce minor inconsistencies in structural representation.
Replacing $\mathcal{L}{2}$ with $\mathcal{L}{1}$ leads to a drop in PSNR, SSIM, and MVSC, and a higher FID of 8.09.
This suggests that L2 loss better stabilizes optimization, particularly for the Transformer-based encoder and decoder in BEV-VAE.
With the full loss combination $\mathcal{L}_{\mathrm{G}}$, PSNR and SSIM remain high, MVSC is well-preserved, and FID reaches its lowest value of 5.10. 
This demonstrates that the complete loss design effectively balances geometric consistency and visual fidelity, significantly improving reconstruction realism.

%% file: tabs/rec_all.tex
\begin{tabular}{ccccccc}
\toprule
\textbf{Model} & \textbf{Latent Shape} & \textbf{Training Data} & \textbf{PSNR$\uparrow$} & \textbf{SSIM$\uparrow$} & \textbf{MVSC$\uparrow$} & \textbf{FID$\downarrow$} \\
\midrule
SD-VAE & $32\times32\times4$  & 5.85B images         & \textbf{29.63} & \textbf{0.8283} & \textbf{0.9292} & \textbf{2.18} \\
\midrule
\textbf{BEV-VAE} & $32\times32\times4$  & 155K × 6 views & 23.48 & 0.6039 & 0.8994 & 17.83 \\
\textbf{BEV-VAE} & $32\times32\times8$  & 155K × 6 views & 24.53 & 0.6569 & 0.9107 & 13.08 \\
\textbf{BEV-VAE} & $32\times32\times16$ & 155K × 6 views & 25.73 & 0.7124 & 0.9222 & 11.42 \\
\textbf{BEV-VAE} & $32\times32\times32$ & 155K × 6 views & 26.32 & 0.7455 & 0.9291 & 13.72 \\
\bottomrule
\end{tabular}
\vspace{1mm}
\caption*{(a) Reconstruction metrics on nuScenes across different dimensions, with SD-VAE as reference.}
\vspace{-2mm}
% --------------------- 子表 b：AV2 ---------------------
\begin{tabular}{ccccccc}
\toprule
\textbf{Model} & \textbf{Latent Shape} & \textbf{Training Data} & \textbf{PSNR$\uparrow$} & \textbf{SSIM$\uparrow$} & \textbf{MVSC$\uparrow$} & \textbf{FID$\downarrow$} \\
\midrule
SD-VAE & $32\times32\times4$  & 5.85B images         & \textbf{27.81} & \textbf{0.8229} & 0.8962 & \textbf{1.87} \\
\midrule
\textbf{BEV-VAE} & $32\times32\times4$  & 224K × 7 views  & 22.99 & 0.6318 & 0.8270 & 7.47  \\
\textbf{BEV-VAE} & $32\times32\times8$  & 224K × 7 views  & 24.02 & 0.6870 & 0.8827 & 5.10  \\
\textbf{BEV-VAE} & $32\times32\times16$  & 224K × 7 views & 25.49 & 0.7529 & 0.9226 & 3.62 \\
\textbf{BEV-VAE} & $32\times32\times32$  & 224K × 7 views & 26.68 & 0.8004 & \textbf{0.9505} & 3.02 \\
\bottomrule
\end{tabular}
\vspace{1mm}
\caption*{(b) Reconstruction metrics on AV2 across different dimensions, with SD-VAE as reference.}

%% file: tabs/gen_nusc.tex
\begin{tabular}{ccccc}
\toprule
Scale & $d=4$ & $d=8$ & $d=16$ & $d=32$ \\
\midrule
$s=0$ & 30.04 & 32.01 & 40.31 & 47.10 \\
$s=1$ & 27.19 & 24.17 & 30.03 & 38.58 \\
$s=2$ & 24.14 & 22.43 & 24.63 & 32.78 \\
$s=3$ & 23.41 & 22.22 & 23.45 & 31.03 \\
$s=4$ & 22.94 & 22.11 & 23.42 & 30.31 \\
$s=5$ & 23.06 & \textbf{21.14} & 23.74 & 30.52 \\
\bottomrule
\end{tabular}

%% file: tabs/gen_av2.tex
\begin{tabular}{ccccc}
\toprule
Scale & $d=4$ & $d=8$ & $d=16$ & $d=32$ \\
\midrule
$s=0$ & 23.63 & 30.05 & 33.12 & 32.66 \\
$s=1$ & 16.87 & 15.28 & 23.28 & 21.82 \\
$s=2$ & 13.48 & 11.15 & 16.65 & 16.14 \\
$s=3$ & 12.72 & \textbf{10.68} & 14.96 & 15.31 \\
$s=4$ & 13.03 & 11.06 & 14.73 & 15.97 \\
$s=5$ & 13.34 & 11.92 & 15.15 & 17.01 \\
\bottomrule
\end{tabular}

%% file: tabs/benchmark.tex
\begin{tabular}{cccc}
\toprule
Method & Paradigm & Extra Prior & FID$\downarrow$ \\
\midrule
BEVGen & Autoregression & None & 25.54\\ 
Panacea & Diffusion  &  Stable Diffusion & 16.96\\ 
MagicDrive & Diffusion  &  Stable Diffusion & 16.20\\ 
DrivingDiffusion & Diffusion  &  Stable Diffusion & 15.83\\ 
DriveWM & Diffusion  &  Stable Diffusion & 12.99\\ 
\midrule
\textbf{\method~w/ DiT}  & Diffusion & None & 21.14\\ 
\bottomrule 
\end{tabular}

%% file: tabs/ablation.tex
\begin{tabular}{lcccc}
\toprule
Setting  & PSNR$\uparrow$ & SSIM$\uparrow$ & MVSC$\uparrow$ & FID$\downarrow$ \\
\midrule
wo/ $\mathcal{L}_{\mathrm{perceptual}}$   & 25.19 & 0.7203 & 0.8698 & 68.99 \\
wo/ $\mathcal{L}_{\mathrm{A}}$  & 24.87 & 0.7135 & 0.9053 & 13.62  \\
w/ $\mathcal{L}_{1}$ instead of $\mathcal{L}_{2}$  & 23.41 & 0.6780 & 0.8632 & 8.09 \\
$\mathcal{L}_{\mathrm{G}}$  & 24.02 & 0.6870 & 0.8827 & 5.10 \\
\bottomrule
\end{tabular}

%% file: secs/5_conclusion.tex
\section{Conclusion}
This paper proposes BEV-VAE, a novel framework for multi-view image generation in autonomous driving. It encodes multi-view images into a compact BEV latent space and performs diffusion-based generation using a DiT. 
Experiments on nuScenes and AV2 validate the effectiveness of BEV-VAE, which achieves competitive performance on nuScenes and scales well to AV2.
Further analysis explores the impact of latent dimensions and guidance scale, while qualitative results highlight its controllable view synthesis through camera and object manipulations.
As a next step, future work could explore temporal modeling for dynamic scenes, integrate physical priors for enhanced consistency, and investigate downstream applications in motion prediction and planning. Overall, BEV-VAE bridges generative modeling and 3D scene understanding, offering a scalable and structured approach to multi-view image generation in autonomous driving.

%% file: supp.tex
\begin{center}
  \textbf{\Large Supplementary Material for BEV-VAE}
\hspace{1cm}
\end{center}
\appendix
The supplementary material offers additional context and results that enhance the main paper on BEV-VAE. 
First, Sec.~\ref{sec:preliminary} provides the core principles of the generative models used in our framework.
Then, Sec.~\ref{sec:evaluation} then explains the multi-view spatial consistency (MVSC) metric in detail and compares it with prior methods. In Sec.~\ref{sec:reconstruction}, we provide further qualitative results on multi-view reconstruction, including renderings from varied camera poses. 
Then, Sec.~\ref{sec:generation} presents examples of fine-grained 3D object layout control, enabling adjustments in the number, position, and orientation of vehicles. 
Lastly, Sec.~\ref{sec:limitations} discusses limitations related to resolution and the need for large-scale training data.
\section{Preliminary for Generative Models}
\label{sec:preliminary}
\noindent\textbf{VAE} is trained by maximizing the Evidence Lower Bound (ELBO) as follows:
\begin{equation}
\log p_{\theta}(x) \geq \mathbb{E}_{q_{\phi}(z|x)} \left[ \log p_{\theta}(x | z) \right] - D_{\mathrm{KL}} \left( q_{\phi}(z | x) \, \| \, p_{\theta}(z) \right),
\end{equation}
where $x$ is the input data, $z$ is the latent variable, $\phi$ and $\theta$ are the encoder and decoder parameters, respectively. 
The first term ensures that the decoder $p_\theta(x \mid z)$ can accurately reconstruct $x$ from the latent variable $z$, and the second term penalizes the divergence between the posterior $q_\phi(z \mid x)$ and the prior $p(z)$, typically $\mathcal{N}(0, I)$, encouraging a structured and continuous latent space.

\noindent\textbf{Diffusion models} define a forward process that gradually adds Gaussian noise to real data $x_0$, formulated as:
\begin{equation}
q(x_t \mid x_0)=\mathcal{N}(x_t ; \sqrt{\bar{\alpha}_t} x_0,(1-\bar{\alpha}_t) \mathbf{I}),
\end{equation}
where $\bar{\alpha}_t$ are pre-defined noise scheduling coefficients, enabling direct sampling of $x_t$ from $x_0$ without iterative noise application. With reparameterization, the noised sample is:
\begin{equation}
x_t=\sqrt{\bar{\alpha}_t} x_0+\sqrt{1-\bar{\alpha}_t} \epsilon_t, \quad \epsilon_t \sim \mathcal{N}(0, \mathbf{I}).
\end{equation}
This highlights the relationship between $x_0$ and noise $\epsilon_t$, enabling training via noise prediction.
The reverse process learns to iteratively denoise $x_t$ back to $x_0$, where
\begin{equation}
p_\theta(x_{t-1} \mid x_t)=\mathcal{N}(x_{t-1} ; \mu_\theta(x_t), \sigma_t^2\mathbf{I}),
\end{equation}
The mean $\mu_\theta(x_t)$ is predicted by the model, while the variance $\sigma_t^2$ is fixed as in DDPM. The ELBO is minimized during training, simplifying to a noise prediction objective:
\begin{equation}
\mathcal{L}_{\text{simple}}(\theta)=\mathbb{E}[\|\epsilon_\theta(x_t)-\epsilon_t\|_2^2].
\end{equation}
Sampling starts from a standard Gaussian $x_T \sim \mathcal{N}(0, \mathbf{I})$ and iteratively denoises via $p_\theta(x_{t-1} \mid x_t)$ to generate samples consistent with the target distribution.

\noindent\textbf{Classifier-Free Guidance (CFG)} enhances conditional diffusion models by adjusting the sampling process to prioritize samples with high $p(c \mid x)$. By applying Bayes' rule, the gradient formulation is derived as:
\begin{equation}
\nabla_x \log p(c \mid x)=\nabla_x \log p(x \mid c)-\nabla_x \log p(x),
\end{equation}
which implies that increasing $p(c \mid x)$ can be achieved by adjusting the diffusion trajectory toward higher $p(x \mid c)$. The reverse diffusion process follows:
\begin{equation}
p_\theta(x_{t-1} \mid x_t, c)=\mathcal{N}(x_{t-1} \mid \mu_\theta(x_t, c), \sigma_t^2\mathbf{I}).
\end{equation}
To guide the diffusion towards the conditional distribution, CFG modifies the noise prediction as:
\begin{equation}
\hat{\epsilon}_\theta(x_t, c)=\epsilon_\theta(x_t, \emptyset)+s \cdot(\epsilon_\theta(x_t, c)-\epsilon_\theta(x_t, \emptyset)) \propto \epsilon_\theta(x_t, \emptyset)+s \cdot \nabla_x \log p(c \mid x_t).
\end{equation}
During training, conditioning is randomly dropped to learn both conditional and unconditional noise predictions.

\section{Evaluation with Multi-View Spatial Consistency}
\begin{figure}[t]
    \centering
    \includegraphics[width=1.0\linewidth]{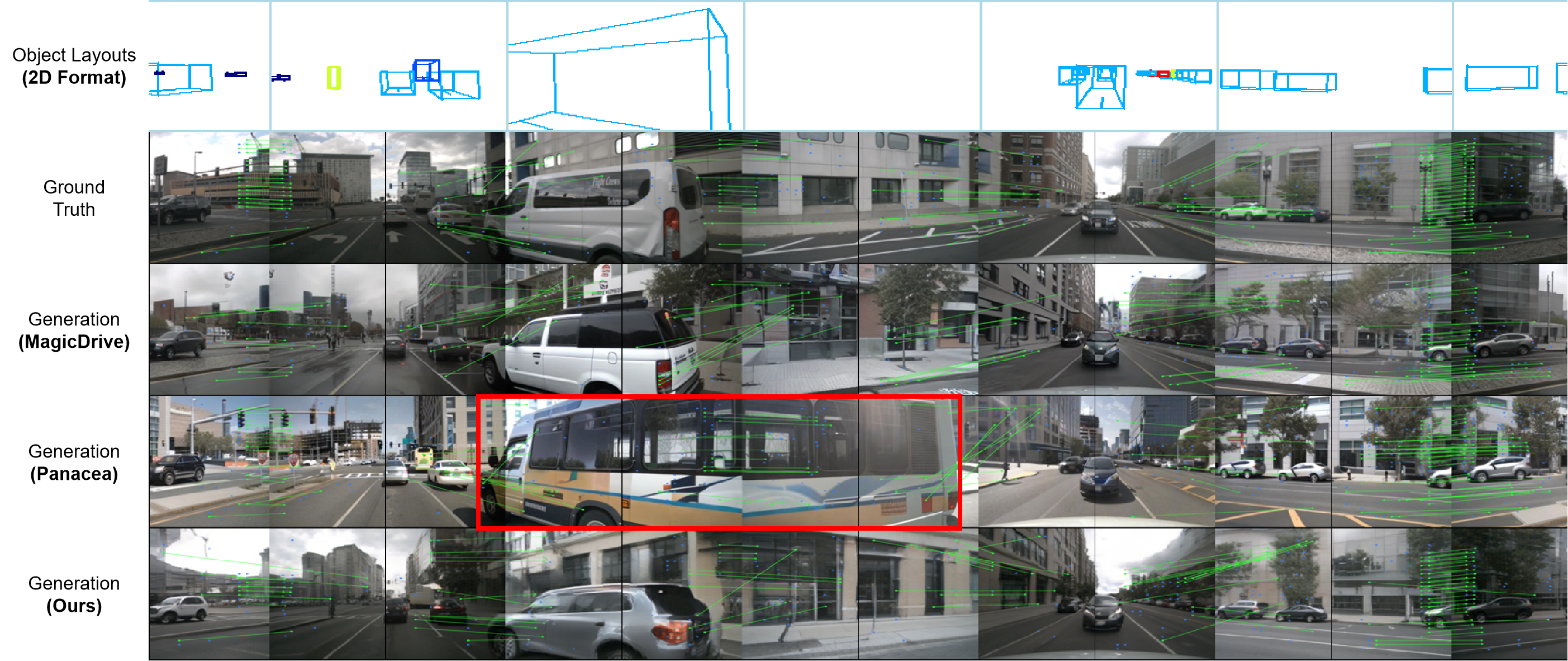}
    \vspace{-4mm}
    \caption{\textbf{Multi-View Spatial Consistency (MVSC) on nuScenes.} The comparison is based on images generated by different methods. Row 1 shows the projections of 3D object layouts onto the image plane. Row 2 presents the corresponding validation images. Rows 3–5 display the results generated by MagicDrive, Panacea, and our method, respectively. To better visualize spatial consistency across adjacent views, each row of images is shifted to the right by half an image width. Vertical black lines mark the centerlines of each camera view. Red boxes indicate regions where the generated vehicles are significantly misaligned with the ground-truth layouts.}
    \label{fig:mvsc}
    \vspace{-4mm}
\end{figure}

Evaluating images with pre-trained models is a common practice, with metrics such as Inception Score (IS), Fréchet Inception Distance (FID), and Learned Perceptual Image Patch Similarity (LPIPS) widely used. To assess spatial consistency in multi-view generation, a matching-based metric is introduced.
Following prior works such as BEVGen and DriveWM, a pre-trained LoFTR model is employed to perform keypoint matching between adjacent views. Given that the overlapping regions between adjacent views typically cover no more than half of the image centered horizontally, each image is divided vertically into left and right halves. For each adjacent camera pair, keypoint matching is performed between the two bordering half-images, as shown in Fig.~\ref{fig:mvsc}.
The proposed Multi-View Spatial Confidence (MVSC) is then defined as the ratio of this average confidence from reconstructed or generated images to that from real images, serving as an indicator of spatial consistency across views.

Based on the same MVSC metric, Table~\ref{tab:mvsc} compares MagicDrive, Panacea, and our method. Although our approach yields a higher FID on nuScenes compared to prior methods, it achieves better spatial consistency than MagicDrive. While Panacea reports a higher MVSC score, this advantage comes partly from leveraging more control signals, such as BEV maps and object depth images. Moreover, as shown in the red box of Fig.~\ref{fig:mvsc}, Panacea generates vehicles that significantly deviate from the ground-truth 3D bounding boxes, which may result from the distortion introduced by perspective projection and cross-view attention mechanisms.

In contrast, BEV-VAE adopts a more straightforward and physically grounded representation of object layouts. MagicDrive encodes 3D boxes using Fourier embeddings and MLPs, which are then fused with image features via cross-attention. Panacea projects 3D boxes into the image plane and aligns them at the pixel level using ControlNet. In our case, object layouts are represented as binary occupancy maps directly in the BEV space, inherently aligned with the BEV representation in 3D without requiring any additional projection or alignment process.
Camera poses are also utilized in a physically consistent manner. By rotating the extrinsic matrix applied to the BEV representation, new views can be rendered directly. This 3D-to-2D mapping ensures that spatial relationships are preserved across views, resulting in inherently consistent multi-view generation.
\label{sec:evaluation}
\begin{table}[t]
\centering
% \vspace{-4mm}
\setlength{\tabcolsep}{5.0pt}
\scalebox{0.9}{\input{tabs/mvsc_compare}}
% \vspace{-6pt}
\caption{\textbf{Comparison on nuScenes: image quality, spatial consistency, and conditions}}
\label{tab:mvsc}
\vspace{-4mm}
\end{table}
\section{Reconstruction with Camera Pose Control}
\label{sec:reconstruction}
To demonstrate that the BEV latent space possesses both 3D structure and complete semantic information, we reconstruct multi-view images from BEV representations under systematically rotated camera extrinsics.
As shown in~\cref{fig:rec_e1_nusc,fig:rec_e1_av2,fig:rec_e2_nusc,fig:rec_e2_av2,fig:rec_e3_nusc,fig:rec_e3_av2}, Row 1 presents the validation images, while Rows 2–8 show reconstructed multi-view images with all camera extrinsics rotated by 15$^\circ$, 10$^\circ$, 5$^\circ$, 0$^\circ$, -5$^\circ$, -10$^\circ$, and -15$^\circ$, respectively. 
This showcases the capability of the BEV latent space to synthesize novel views by manipulating camera poses. To highlight the effect of view synthesis, the latent dimension is set to 32.

\section{Generation with Precise 3D Object Control}
\label{sec:generation}
To demonstrate that the BEV latent space supports precise control based on structured 3D object layouts, we generate multi-view images by selectively removing different vehicles from the same scene.
As shown in Fig.~\ref{fig:gen_e1_nusc} and Fig.~\ref{fig:gen_e1_av2}, Row 1 presents real images from the validation set, and Row 2 shows the reconstructed images. Row 3 displays images generated from the corresponding 3D bounding boxes. 
Rows 4–8 further illustrate controllable generation by selectively removing specific vehicles from the input layouts, with the removed objects indicated by numerical labels. 
In addition, Fig.~\ref{fig:gen_rot_box} demonstrates that the orientation of a vehicle in the generated images can be precisely controlled by rotating its 3D bounding box within the same scene layout.
It is worth noting that the same 3D bounding box may lead to different object appearances across generated views.

\section{Limitations in Resolution and Data Scale}
\label{sec:limitations}
Our framework is fully based on Transformer architectures and has been validated at a resolution of $256{\times}256$, demonstrating the feasibility of this design paradigm. However, compared to methods that fine-tune large pre-trained diffusion models (e.g., Stable Diffusion), our generated and reconstructed images tend to appear blurrier—particularly on the nuScenes dataset. This is primarily due to the lack of pre-trained image priors and the relatively low resolution used during training, rather than limitations in model capacity.

\begin{figure}[t]
    \centering
    \includegraphics[width=1.0\linewidth]{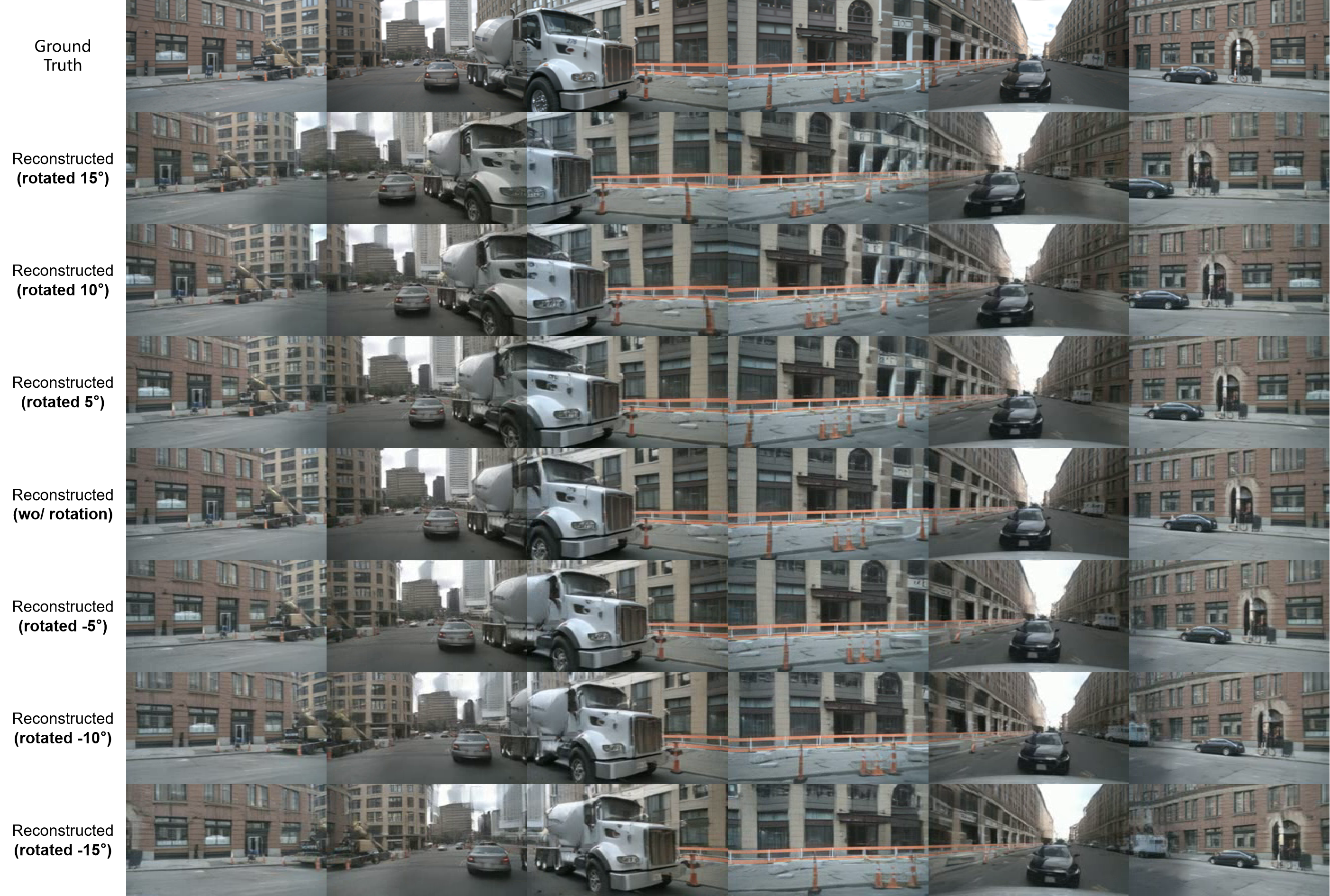}
    \vspace{-4mm}
    \caption{\textbf{Example 1 on nuScenes: Novel view synthesis via camera pose modifications.}}
    \label{fig:rec_e1_nusc}
    \vspace{-4mm}
\end{figure}

\begin{figure}[t]
    \centering
    \includegraphics[width=1.0\linewidth]{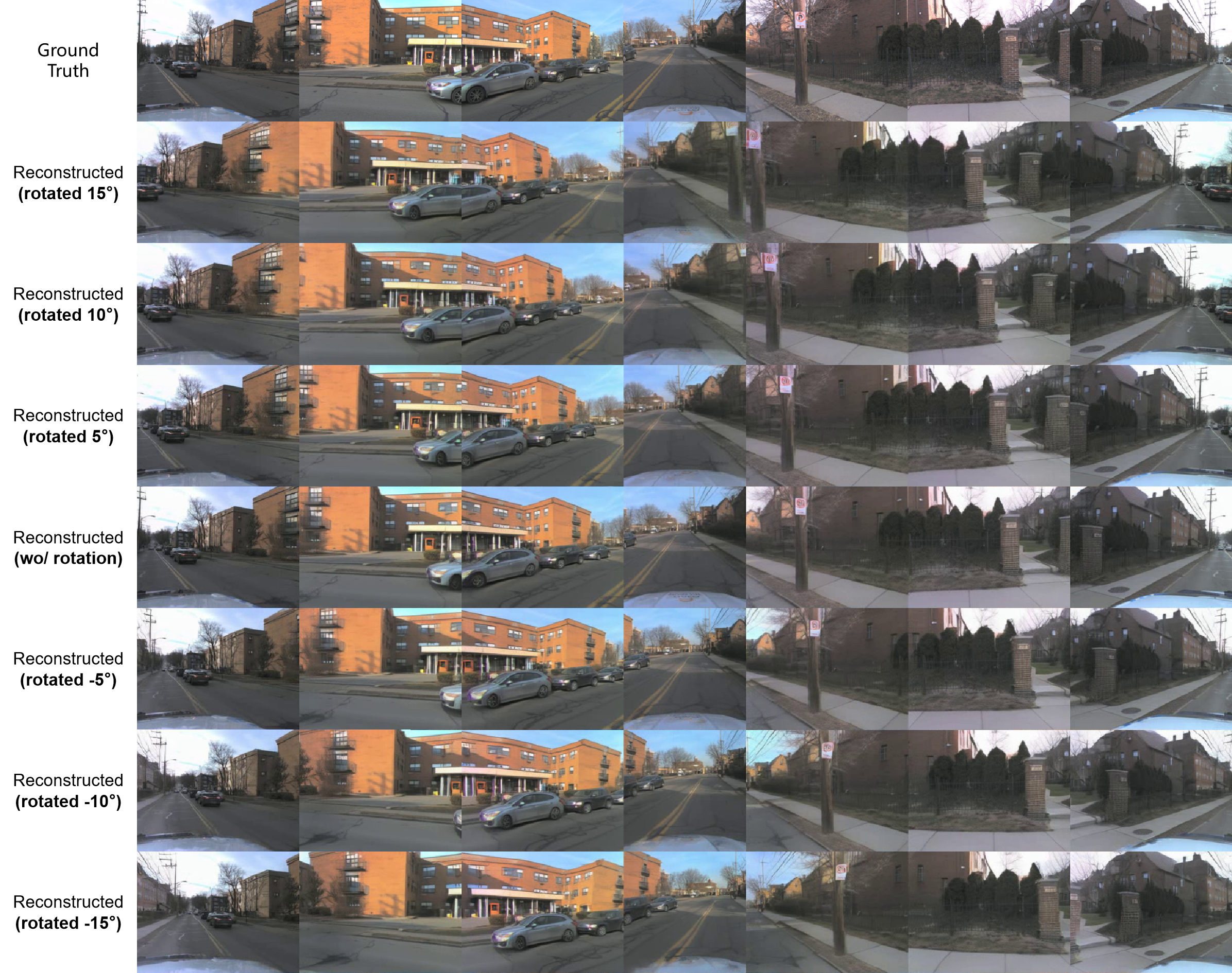}
    \vspace{-4mm}
    \caption{\textbf{Example 1 on AV2: Novel view synthesis via camera pose modifications.}}
    \label{fig:rec_e1_av2}
    \vspace{-4mm}
\end{figure}

\begin{figure}[t]
    \centering
    \includegraphics[width=1.0\linewidth]{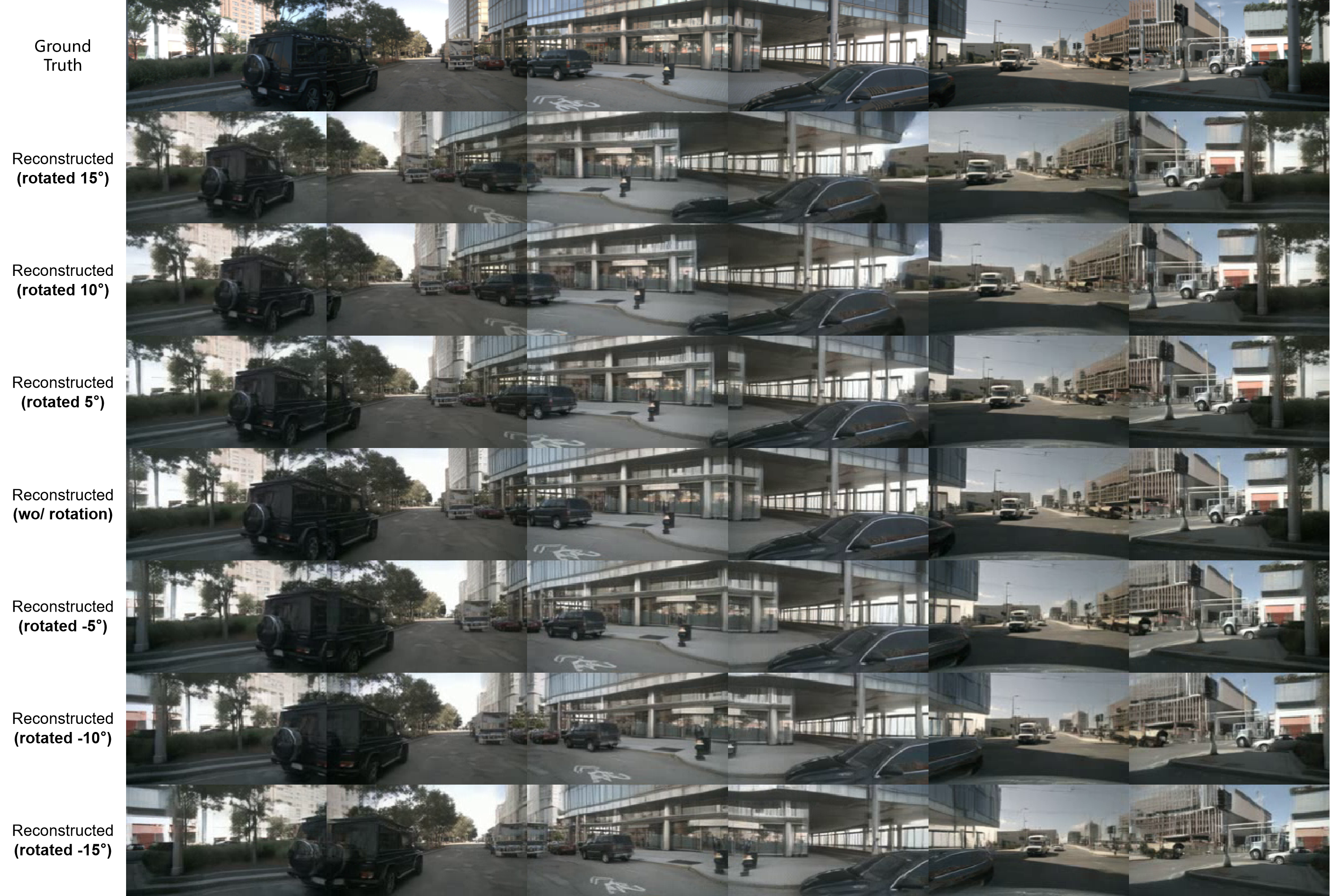}
    \vspace{-4mm}
    \caption{\textbf{Example 2 on nuScenes: Novel view synthesis via camera pose modifications.}}
    \label{fig:rec_e2_nusc}
    \vspace{-4mm}
\end{figure}

\begin{figure}[t]
    \centering
    \includegraphics[width=1.0\linewidth]{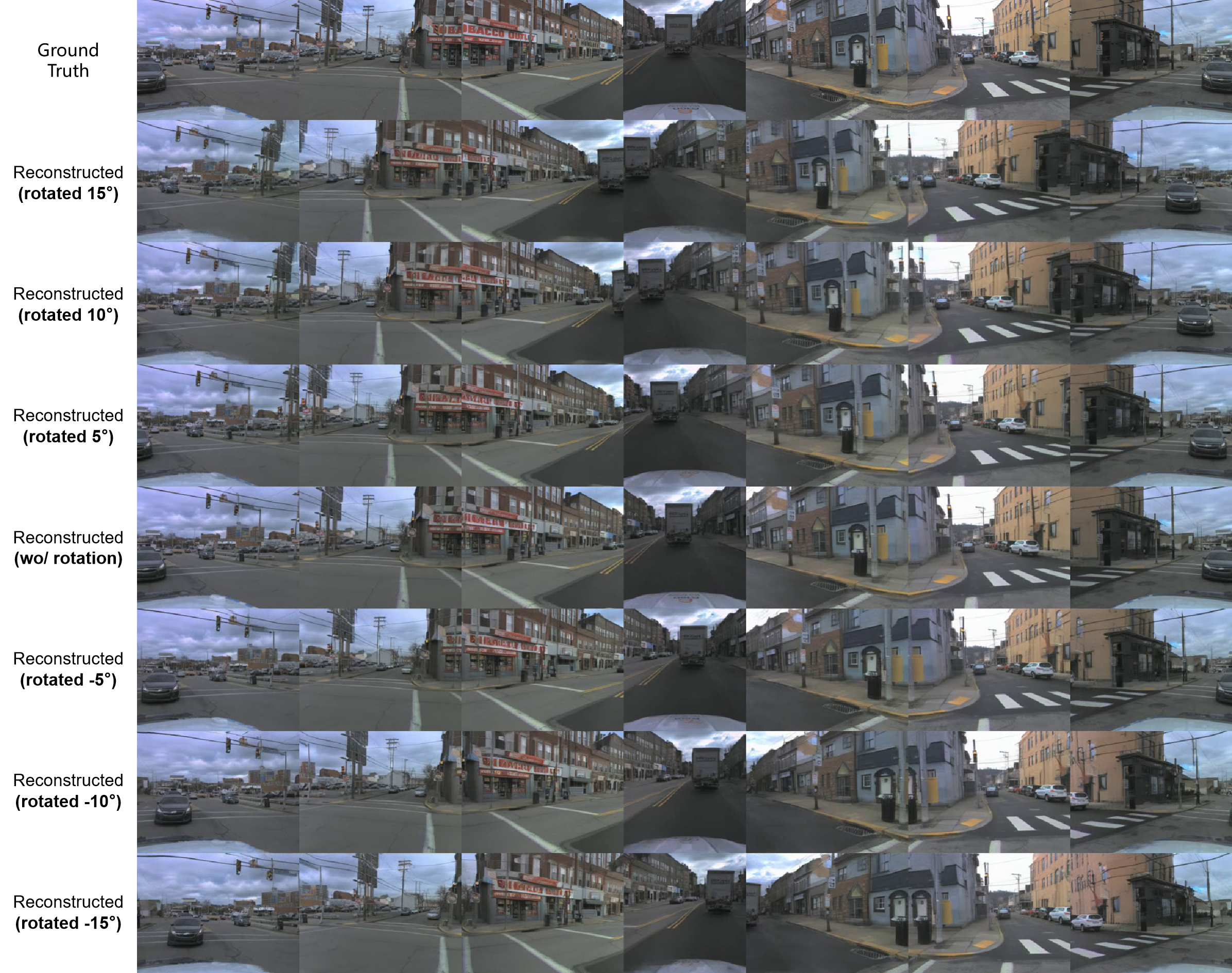}
    \vspace{-4mm}
    \caption{\textbf{Example 2 on AV2: Novel view synthesis via camera pose modifications.}}
    \label{fig:rec_e2_av2}
    \vspace{-4mm}
\end{figure}

\begin{figure}[t]
    \centering
    \includegraphics[width=1.0\linewidth]{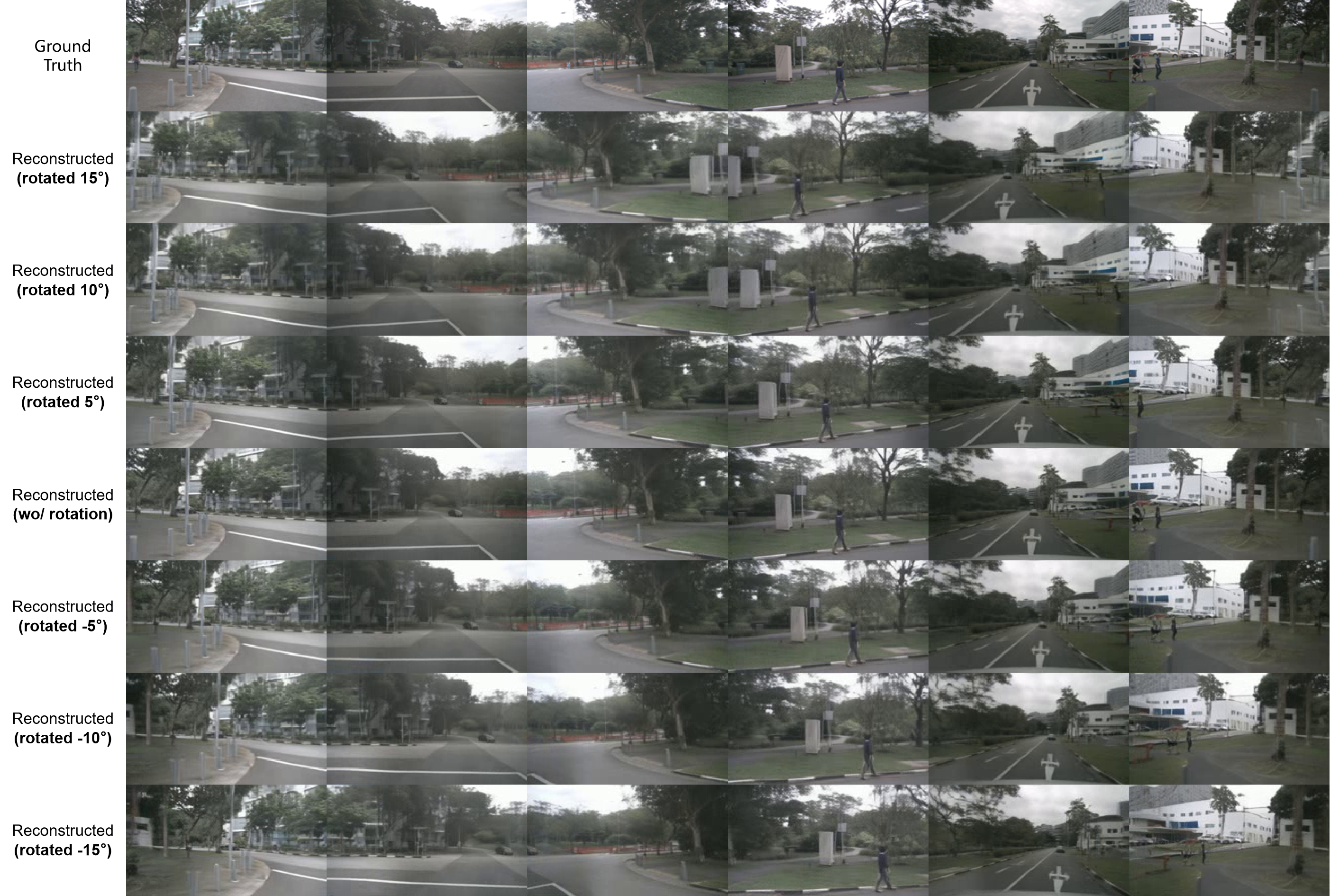}
    \vspace{-4mm}
    \caption{\textbf{Example 3 on nuScenes: Novel view synthesis via camera pose modifications.}}
    \label{fig:rec_e3_nusc}
    \vspace{-4mm}
\end{figure}

\begin{figure}[t]
    \centering
    \includegraphics[width=1.0\linewidth]{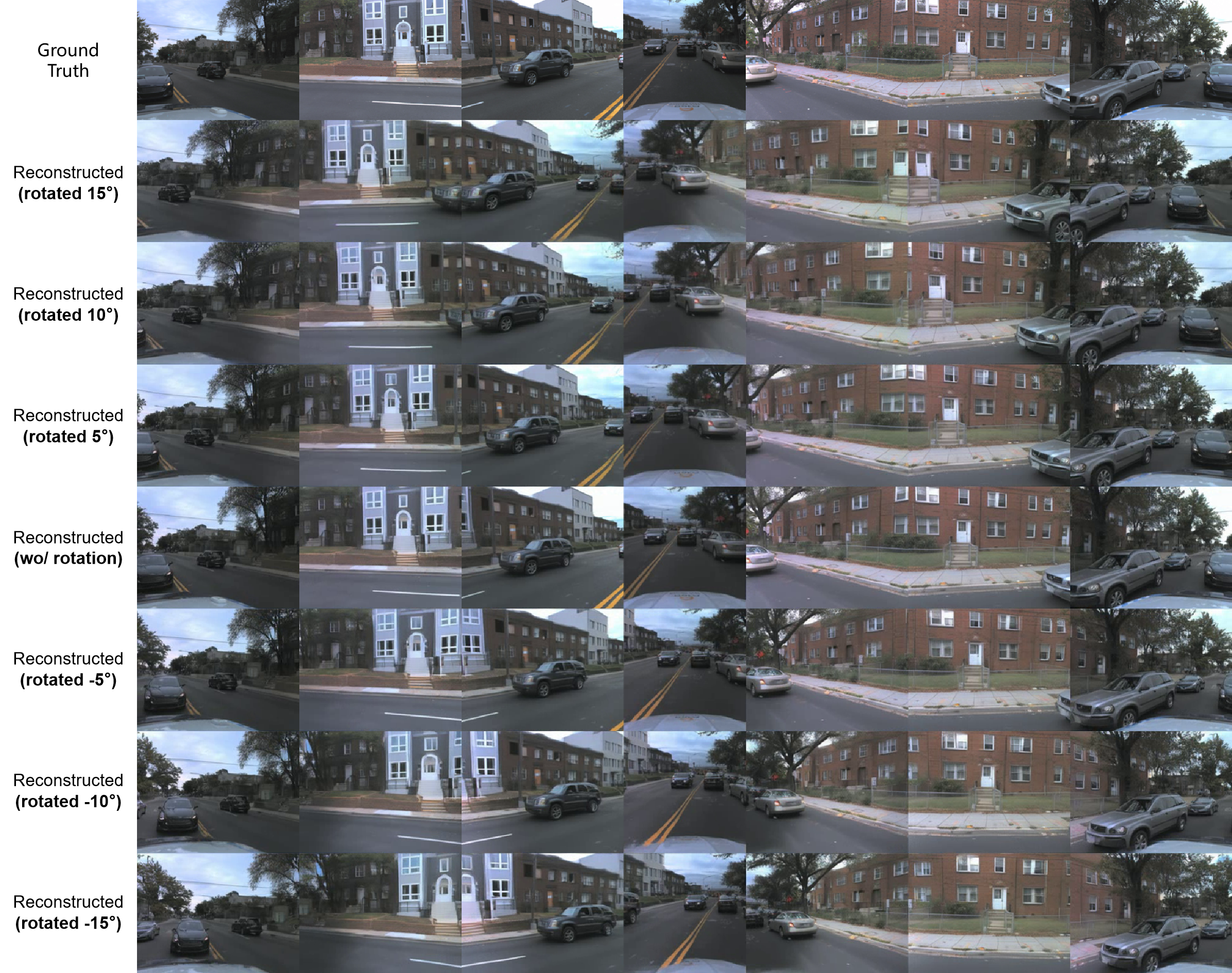}
    \vspace{-4mm}
    \caption{\textbf{Example 3 on AV2: Novel view synthesis via camera pose modifications.}}
    \label{fig:rec_e3_av2}
    \vspace{-4mm}
\end{figure}

\begin{figure}[t]
    \centering
    \includegraphics[width=1.0\linewidth]{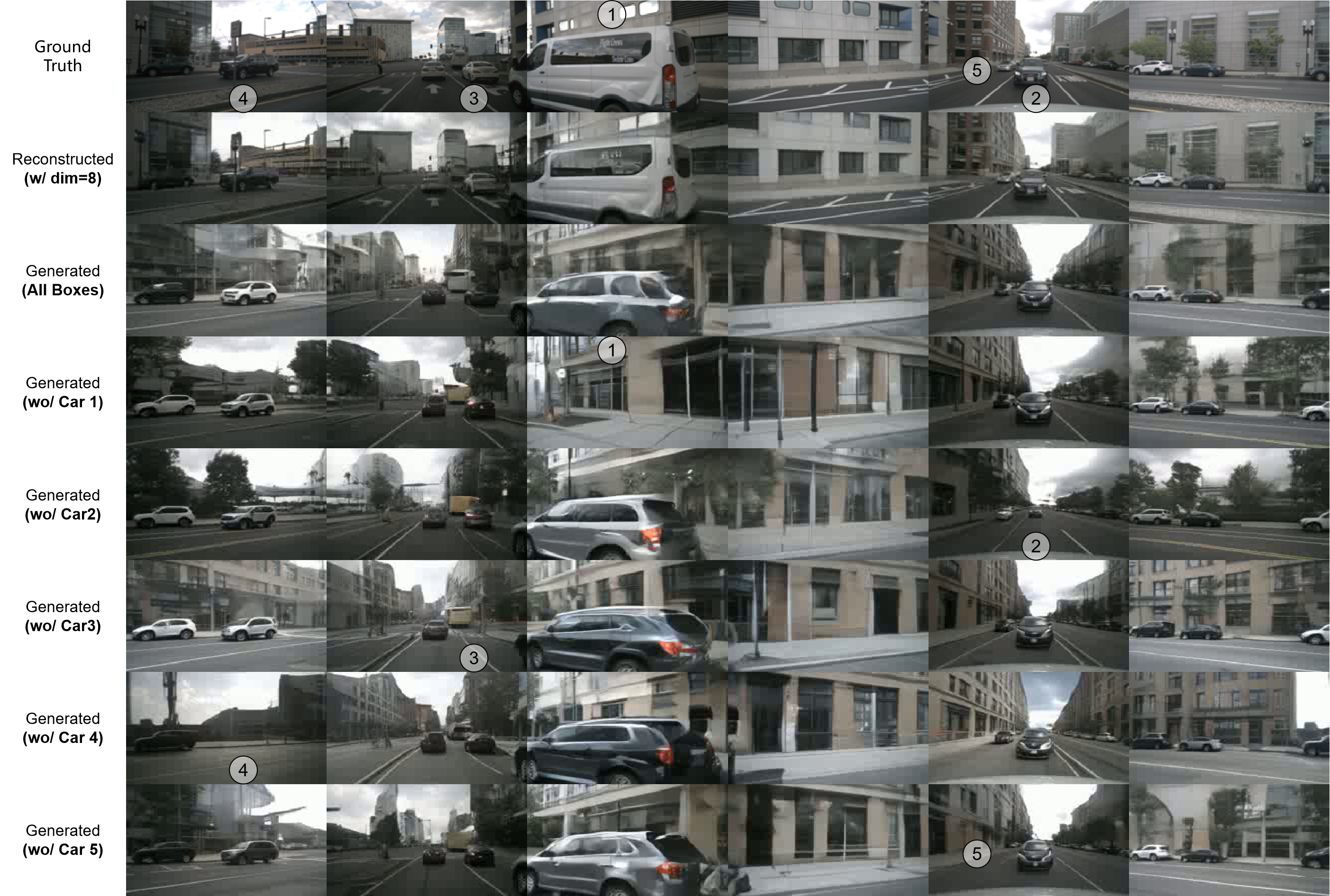}
    \vspace{-6mm}
    \caption{\textbf{Multi-view image generation on nuScenes with 3D object layout editing.}}
    \label{fig:gen_e1_nusc}
    \vspace{-4mm}
\end{figure}

\begin{figure}[t]
    \centering
    \includegraphics[width=1.0\linewidth]{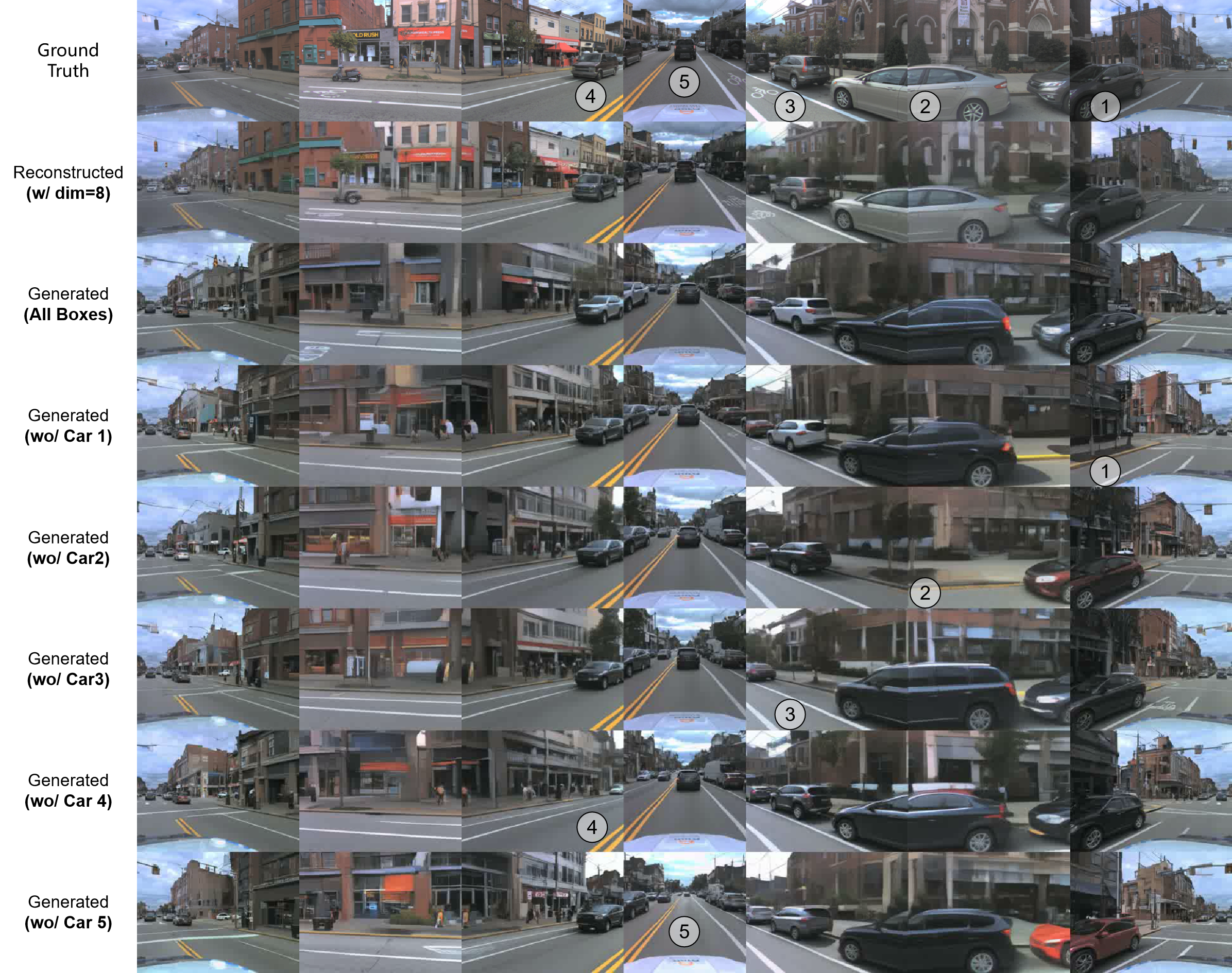}
    \vspace{-6mm}
    \caption{\textbf{Multi-view image generation on AV2 with 3D object layout editing.}}
    \label{fig:gen_e1_av2}
    \vspace{-4mm}
\end{figure}

Another critical factor is dataset scale. Argoverse 2 (AV2) contains approximately 1.5$\times$ more training data than nuScenes, and this difference is clearly reflected in the results. As shown in~\cref{fig:rec_e1_nusc,fig:rec_e1_av2,fig:rec_e2_nusc,fig:rec_e2_av2,fig:rec_e3_nusc,fig:rec_e3_av2,fig:gen_e1_nusc,fig:gen_e1_av2}, both reconstruction and generation on AV2 outperform those on nuScenes by a notable margin. To the best of our knowledge, our approach is the first to support generation from 7 surround-view cameras on AV2, and thus no prior baseline exists for direct comparison.
This progression from nuScenes to AV2 highlights the scaling potential of our method. BEV-VAE fundamentally learns a generalizable 2D-to-3D encoding and 3D-to-2D decoding process. Unlike direct image generation methods, our framework requires sufficient data to capture the underlying spatial structure and to ensure consistent multi-view generation through a structured BEV latent space.

\clearpage
\vspace*{-5mm}  
\noindent
\begin{minipage}{\linewidth}
    \centering
    \includegraphics[width=1.0\linewidth]{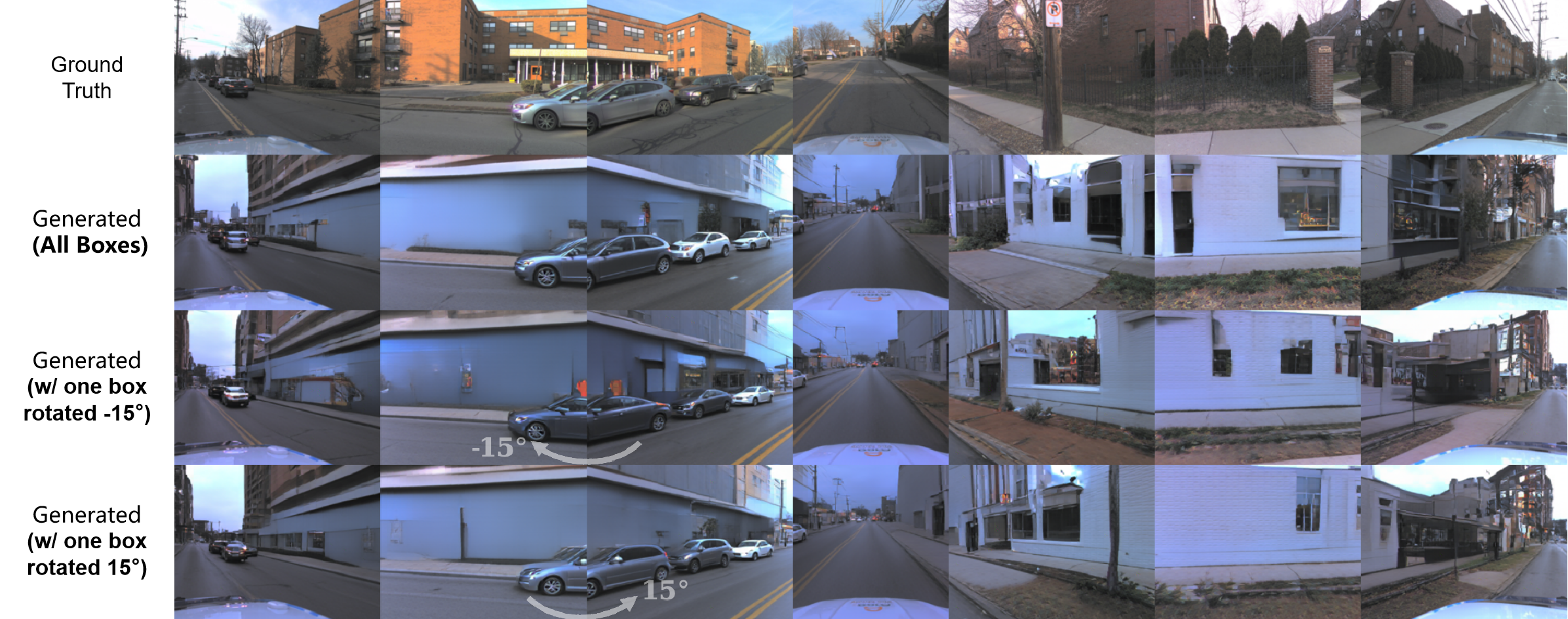}
    \vspace{-6mm}
    \captionof{figure}{\textbf{Rotating the orientation of a specific vehicle on AV2.} Row 1 presents validation images and Row 2 shows generated images. Rows 3 and 4 depict the same vehicle rotated 15° clockwise and counterclockwise on the ego vehicle's horizontal plane.}
    \label{fig:gen_rot_box}
\end{minipage}
\vspace{-4mm}

%% file: tabs/mvsc_compare.tex
\begin{tabular}{cccccc}
\toprule
Method                   & FID$\downarrow$ & MVSC$\uparrow$  & Object Layouts & Camera Poses & Other Conditions\\
\midrule
MagicDrive               & 16.20           & 0.8310          & Fourier embedding(1D) & Fourier embedding & Text, map.  \\ 
Panacea                  & 16.96           & 0.9189          & Perspective projection (2D) & Pseudo-color image & Text, map, depth. \\ 
\midrule
\textbf{Ours}  & 21.14           & 0.8902          & Binary occupancy (3D) & Extrinsic matrix & None \\ 
\bottomrule 
\end{tabular}